\documentclass[sigconf]{acmart}
\makeatletter
\let\@authorsaddresses\@empty
\makeatother

\AtBeginDocument{%
  }

\copyrightyear{2025}
\acmYear{2025}
\setcopyright{acmlicensed}\acmConference[SA Conference Papers '25]{SIGGRAPH Asia 2025 Conference Papers}{December 15--18, 2025}{Hong Kong, Hong Kong}
\acmBooktitle{SIGGRAPH Asia 2025 Conference Papers (SA Conference Papers '25), December 15--18, 2025, Hong Kong, Hong Kong}
\acmDOI{10.1145/3757377.3763816}
\acmISBN{979-8-4007-2137-3/2025/12}

\usepackage{enumitem}
\usepackage{multirow}
\def\ie{\textit{i.e.}}
\def\eg{\textit{e.g.}}
\def\etc{\textit{etc.}}

\newcommand{\dataset}{\textit{V3DBench} }
\newcommand{\name}{{\textit{\textbf{Shape-for-Motion}}}}

\newcommand{\figref}[1]{Fig.~\ref{#1}}
\newcommand{\reqref}[1]{Eq.~\ref{#1}}
\newcommand{\secref}[1]{Sec.~\ref{#1}}
\newcommand{\tableref}[1]{Tab.~\ref{#1}}

\makeatletter
\def\@fnsymbol#1{\ensuremath{%
  \ifcase#1\or \dagger\or \ddagger\or \mathsection\or \mathparagraph\or \|\or
  **\or \dagger\dagger\or \ddagger\ddagger \else\@ctrerr\fi}}
\makeatother

\begin{document}

\title{Shape-for-Motion: Precise and Consistent Video Editing With 3D Proxy}

\author{Yuhao Liu}
\email{yuhliu9-c@my.cityu.edu.hk}
\affiliation{%
  \institution{City University of Hong Kong}
  \country{Hong Kong SAR, China}}

\author{Tengfei Wang}
\email{tfwang@connect.ust.hk}
\authornote{Co-corresponding authors.}
\affiliation{%
  \institution{Tencent}
  \country{China}}

\author{Fang Liu}
\email{fliu66-c@cityu.edu.hk}
\affiliation{%
  \institution{City University of Hong Kong}
  \country{Hong Kong SAR, China}}

\author{Zhenwei Wang}
\email{zhenwwang2-c@my.cityu.edu.hk}
\affiliation{%
  \institution{City University of Hong Kong}
  \country{Hong Kong SAR, China}}

\author{Rynson W.H. Lau}
\email{Rynson.Lau@cityu.edu.hk}
\authornotemark[1]
\affiliation{%
  \institution{City University of Hong Kong}
  \country{Hong Kong SAR, China}}

\renewcommand{\shortauthors}{Yuhao Liu, Tengfei Wang, Fang Liu et al.}



\begin{abstract}
Recent advances in deep generative modeling have unlocked unprecedented opportunities for video synthesis. 
In real-world applications, however, users often seek tools to faithfully realize their creative editing intentions with precise and consistent control. 
Despite the progress achieved by existing methods, ensuring fine-grained alignment with user intentions remains an open and challenging problem. 
In this work, we present Shape-for-Motion, a novel framework that incorporates a 3D proxy for precise and consistent video editing. 
Shape-for-Motion achieves this by converting the target object in the input video to a time-consistent mesh, \ie, a 3D proxy, allowing edits to be performed directly on the proxy and then inferred back to the video frames. 
To simplify the editing process, we design a novel Dual-Propagation Strategy that allows users to perform edits on the 3D mesh of a single frame, and the edits are then automatically propagated to the 3D meshes of the other frames. 
The 3D meshes for different frames are further projected onto the 2D space to produce the edited geometry and texture renderings, which serve as inputs to a decoupled video diffusion model for generating edited results. 
Our framework supports various precise and physically-consistent manipulations across the video frames, including pose editing, rotation, scaling, translation, texture modification, and object composition. 
Our approach marks a key step toward high-quality, controllable video editing workflows. Extensive experiments demonstrate the superiority and effectiveness of our approach.  Project Page: \href{https://shapeformotion.github.io}{\textcolor{magenta}{https://shapeformotion.github.io}}.

\end{abstract}

\begin{teaserfigure}
\centering
\includegraphics[width=0.95\textwidth]{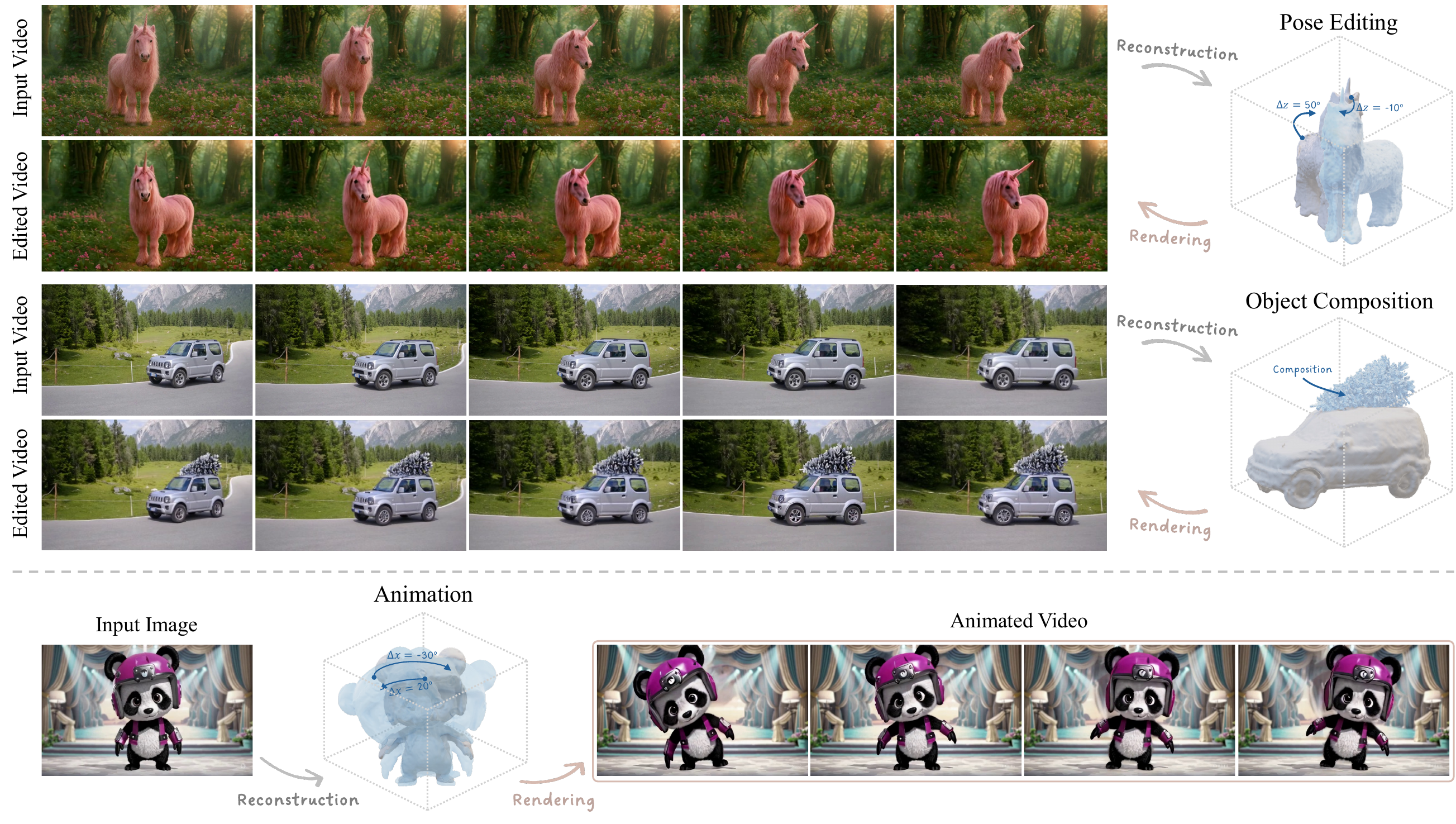}
\vspace{-4mm}
  \caption{The proposed 3D-aware framework, Shape-for-Motion, supports precise and consistent video editing by reconstructing an editable 3D mesh to serve as control signals for video generation. The first two examples demonstrate pose editing (by rotating the back to the right by 50 degrees and the head to the left by 10 degrees) and object composition (by composing a tree from the reference image onto the top of the car). In each example, the first row shows the input video frames, followed by the editing in 3D space at the right end; the bottom row of images shows the corresponding edited frames. In addition, our approach also supports diverse applications, such as Image-to-Video Animation, as shown in the third example.}
  \Description{Nop}
  \label{fig:teaser}
  \vspace{3mm}
\end{teaserfigure}

\maketitle

\section{Introduction}
The rapid proliferation of video content on online platforms, combined with the explosive emergence of video generation models, has fueled the surge in video content generation. 
Recently, controllable video editing, which focuses on modifying the source video content to align with user intentions, has gained increasing attention.

Early approaches \cite{bar2022text2live,wu2023tune} use text as the interaction signal, but text often lacks precision and flexibility for editing spatial attributes. 
Besides text, image-based methods  \cite{ku2024anyv2v, ouyang2024i2vedit} introduce an edited image as guidance and propagate edits across frames, while drag-based methods \cite{deng2025dragvideo,teng2023drag} propose to manually drag anchor points for localized adjustments.
However, they struggle in handling complex editing and ensuring frame-to-frame consistency.

Unlike these existing works, our objective in this work is to develop a video editing system with two key features. 
\textbf{1) Precision.} Accurate controllability gives users precise control over various aspects of the video elements, including object pose, shape, location,  and spatial layout. 
Such fine-grained control often extends to object attributes with quantifiable precision. For example, rotating the panda in the third example of \figref{fig:teaser} by 20 degrees to the left requires precise manipulation. 
\textbf{2) Consistency.} Consistent alignment demands edits to remain coherent across frames. For example, placing a tree on the moving car in the second example of \figref{fig:teaser} requires alignment that simultaneously accounts for the car's motion, rotation, and changing perspectives. 
While these two key features are essential for video editing, it is non-trivial to achieve this with a 2D framework, due to the absence of underlying 3D representations.

To address the above challenges, in this paper, we introduce \name, a novel video editing framework that incorporates a 3D proxy (\ie, mesh) to enable precise editing while maintaining temporal and spatial consistency. 
Our framework follows a 3D-aware workflow by first reconstructing a 3D proxy of the target object from the video, followed by interactive manipulation in the 3D space, and finally, producing a video with the help of the edited 3D proxy.
We leverage three key designs to ensure (1) temporal-consistent 3D proxy reconstruction, (2) consistent 3D editing across frames, and (3) generative rendering from edited 3D to video.

The first is a consistent mesh reconstruction of the target object. 
We note that reconstructing a separate mesh for each frame individually \cite{tang2025lgm} leads to poor consistency due to the lack of inter-frame correspondences.
To address this problem, we propose to reconstruct a consistent mesh representation for the object across all frames using a canonical mesh with a time-varying deformation field. 
However, the limited viewpoint information in a monocular input video often results in unsatisfactory reconstruction. 
To mitigate this, we leverage novel views generated from an existing multi-view generator \cite{voleti2025sv3d} to enhance mesh reconstruction and propose a balanced view sampling strategy to alleviate inconsistencies caused by the generated novel views.

The second design is automatic editing propagation on 3D mesh. 
While editing the mesh in each frame individually suffices for simple global operations, \eg, global rotation or scaling, it becomes impractical for more complex tasks, like localized pose editing as shown in the first example of \figref{fig:teaser}. 
To enable a user-friendly editing process (in which users only need to edit the 3D mesh of a single frame, and the editing is then automatically propagated to other frames), we propose a novel \textit{Dual-Propagation Strategy}, which utilizes learned correspondences between canonical and deformed meshes to propagate geometry and texture edits across frames, enabling the generation of consistent editing.

The final key design is generative rendering from the edited 3D proxy. Once we have edited the object in 3D space, we then need to convert the edited 3D meshes to a high-quality edited video.
However, it is difficult to train such a generation model, due to the absence of paired training data (3D mesh and corresponding video). 
To address this challenge, we adopt a decoupled control strategy in our video diffusion model, treating geometry and texture information from the edited proxy as two separate conditioning signals. 
A self-supervised mixed training strategy is proposed to alternate between geometry and texture controls, enabling the model to preserve its generative capabilities while achieving temporal and spatial consistency across frames. 

\begin{figure*}[t!]
  \includegraphics[width=\textwidth]{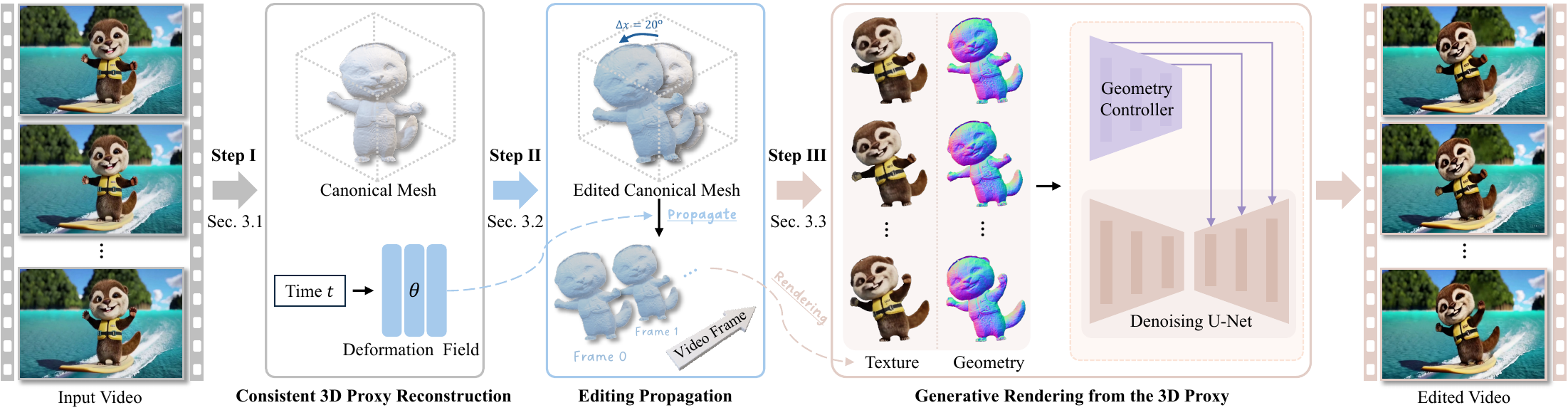}
  \vspace{-7mm}
  \caption{Overview of \name. Our approach is an interactive video editing framework that utilizes an editable 3D proxy (\eg, mesh) to enable users to perform precise and consistent video editing. Given an input video, our approach first converts the target object into a 3D mesh with frame-by-frame correspondences. Users can then perform editing on this 3D proxy once only (\ie, on a single frame), and the edits will be automatically propagated to the 3D meshes of all other frames. The edited 3D meshes are then converted back to 2D geometry and texture maps, which are used as control signals in a decoupled video diffusion model to generate the final edited result. Note that, for better visualization, the colors of the propagated meshes are disabled in Step-$\mathrm{II}$.}
  \label{fig:pipeline}
\end{figure*}

In summary, Shape-for-Motion allows users to perform fine-grained geometry control (\eg, pose editing, rotation, scaling, translation, and object composition) and texture modification in 3D space. 
It also supports appearance editing in 2D space by incorporating 2D editing tools. Extensive experiments and user studies on the new \dataset~dataset demonstrate that our approach generates temporally consistent video edits, and outperforms existing methods qualitatively and quantitatively. Our main contributions are:
\begin{itemize}[noitemsep,topsep=5pt]
    \item We propose \name, a novel video editing framework that incorporates a 3D proxy to allow diverse and precise video object manipulation while ensuring temporal consistency across frames.  
    \item We propose a video object reconstruction method, which produces meshes with correspondences across frames, and a dual propagation strategy. Together, these components enable a user-friendly editing process: users can perform edits directly on the canonical 3D mesh only once, with the edits are automatically propagated to subsequent frames. 
    \item We propose a self-supervised mixed training strategy to train a decoupled video diffusion model that leverages geometry and texture information from the edited 3D proxy as control signals, achieving more consistent editing results.
\end{itemize}

\section{Related Works}

\subsection{Controllable  Video Generation}
Recent success in diffusion models~\cite{ho2020denoising,song2021denoising} have fueled significant progress in image~\cite{rombach2022high,liang2025vodiff} and video generation~\cite{videoworldsimulators2024}. 
Earlier Text-to-Video (T2V)
models \cite{ho2022imagen,blattmann2023stable} are mostly evolved from Text-to-Image (T2I) models by incorporating additional temporal layers. 
Later, Image-to-Video (I2V) methods \cite{guo2023animatediff,guo2024i2v} are proposed to animate an image to produce videos. 
Methods like~\cite{esser2023structure,chen2023control,hu2023videocontrolnet,yang2024direct,huang2025voyager} are also proposed to generate videos by conditioning on a sequence of control frames. 
Several works~\cite{lv2024gpt4motion,shi2024motion,cai2024generative} also explore the application of 3D priors for video generation, either as direct inputs or as intermediate outputs to enhance the generation quality. 
Recently, incorporating video priors for controllable 4D generation~\cite{jiang2024animate3d,bahmani2024tc4d} has also attracted huge research interests.

\subsection{Controllable Video Editing}

To bridge image and video modalities, early methods like Layered Neural Atlases~\cite{kasten2021layered} decompose frames into foreground and background atlases, each with a dedicated map for appearance and geometry, enabling independent object editing and seamless compositing.
In recent years, Control-based methods~\cite{zhang2023controlvideo,ma2023magicstick,wang2022pretraining} are proposed to utilize additional reference maps (\eg, depth or edge maps) to modify object motion and shape. However, creating references for each frame by the user is impractical. 
VideoSwap~\cite{gu2024videoswap} leverages sparse semantic point correspondences to enable shape change in swapped results while aligning with the source motion trajectory.  
First-frame-based methods further reduce this burden by editing on the first frame only and propagating \cite{kagaya2010video} the changes to subsequent frames, either by attention manipulation~\cite{fan2024videoshop,ku2024anyv2v,ceylan2023pix2video} or a fine-tuned model manner~\cite{liu2025generative,mou2024revideo}.
Recently, drag-based methods~\cite{pan2023drag,deng2025dragvideo,teng2023drag} allow users to modify the shape of an object by dragging its local region from a start point to an endpoint. 
Concurrent works, GS-Dit~\cite{bian2025gs} and Diffusion-as-Shader~\cite{gu2025diffusion} represent subjects using point clouds with explicit tracking, offering faster speed without requiring explicit object reconstruction.
The most closely related work, VideoHandles~\cite{koo2025videohandles}, edits 3D object compositions in videos of static scenes with camera motion. In contrast, our method handles dynamic objects with independent motions, enabling precise and consistent video editing.

\subsection{Image Editing Using 3D Proxy}
Recently, a growing body of methods has sought to integrate 3D priors into image editing. 3DIT~\cite{michel2023object} introduces an object-centric image editing framework that edits objects based on language instructions. 
Diffusion Handles~\cite{pandey2024diffusion} edits selected objects by leveraging depth to lift diffusion activations into 3D space. 
The most related works to ours are Image Sculpting \cite{yenphraphai2024image} and 3D-Fixup \cite{cheng20253d}, which use 3D mesh to support precise image editing. In contrast, our approach enables precise and consistent video editing with 3d proxy.

\subsection{Dynamic 3D Reconstruction} 
Compared to NeRF \cite{mildenhall2021nerf,pumarola2021d}, 3D Gaussian Splatting (3DGS) \cite{kerbl20233d} supports both high-quality rendering and real-time efficiency. 
When extended to dynamic scenes, deformable 3DGS incorporates explicit temporal changes of the canonical points \cite{luiten2023dynamic,yang2024deformable,wu20244d}. However, the point-based representation is inherently less suited for precise editing. 
Most recently, DG-Mesh \cite{liu2024dynamic} integrates deformable 3DGS for mesh reconstruction, and can be used to handle dynamic scenes. However, it requires $360^\circ$ capturing of the target object, which is not applicable for general videos.
In contrast, our method aims at reconstructing a consistent 3D mesh from a general monocular video, which typically has limited views of the target object. 
Inspired by multi-view generation~\cite{shi2023mvdream,wang2024phidias}, several methods~\cite{xie2024sv4d,wu2024cat4d,ren2024l4gm} leverage diffusion models to produce novel views for a video, followed by 4D reconstruction. 
However, these methods lack explicit mesh correspondences across frames, limiting their applicability for video editing.
\section{OUR APPROACH}
\label{sec:appraoch}

Given an input video, we aim to enable precise manipulation of the target object. To this end, we propose a novel video editing pipeline (see \figref{fig:pipeline}) consisting of three key steps. 
First, we convert the target object into its 3D mesh to provide a consistent geometry structure for editing (\secref{subsec:recon}). 
Second, we propose a new Dual-Propagation Strategy to enable precise and controllable edits in 3D space, which are then consistently propagated across frames (\secref{subsec:editing}). 
Third, we introduce a self-supervised mixed training strategy for a decoupled video diffusion model, which uses geometry and texture renderings as control signals to produce more consistent results. (\secref{subsec:render}).

\subsection{Consistent 3D Proxy Reconstruction}
\label{subsec:recon}

One naive way to obtain the 3D proxy is to reconstruct each frame individually. However, this separate modeling lacks frame-to-frame correspondences, resulting in poor temporal consistency. 
To address this, we adopt the deformable Gaussian Splatting~\cite{yang2024deformable} to reconstruct the target object in the input video. 
Furthermore, manipulating Gaussian points is not user-friendly and only supports limited editing types. To enable 3D-guided video editing, we aim to obtain consistent 3D meshes as the editing proxy.

\begin{figure}[t!]
\centering
  \includegraphics[width=0.47\textwidth]{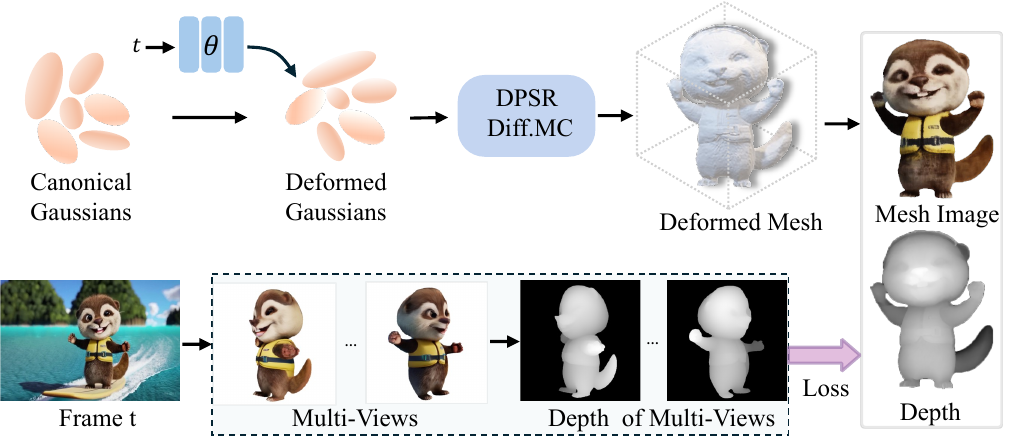}
        \caption{Pipeline of our consistent object reconstruction. We use deformable-3DGS to reconstruct the 3D mesh of the target object in a video by maintaining canonical Gaussians and a time-varying deformation field $\theta$. To achieve complete reconstruction, in addition to the standard Gaussian Splatting (GS) loss, we incorporate multi-view augmentation as additional constraints.}
  \label{fig:stage1}
\end{figure}

\noindent \textbf{Overall Pipeline.} 
The workflow is illustrated in \figref{fig:stage1}. 
At time $t$, we first obtain the deformation of the canonical Gaussians from the deformation field encoded with a learnable MLP $\theta(\cdot)$ to form the deformed Gaussians. 
We then utilize DPSR~\cite{peng2021shape} combined with Marching Cubes (MC) to convert the deformed Gaussians into a Gaussian-propagated deformed mesh. 
To obtain the vertex colors of the deformed mesh, we follow \cite{liu2024dynamic} to store the texture in a canonical color MLP, denoted as $MLP_{c}(\cdot)$. 
By training a backward deformation field $\theta^{-1}(\cdot)$, we map the deformed mesh back to the canonical space at each time $t$, and query the vertex colors via $MLP_{c}(\cdot)$. 
Finally, we perform differentiable rasterization \cite{laine2020modular} to render the depth map, mesh image, and mask of the deformed mesh. Meanwhile, Gaussian image is rendered from the deformed Gaussians.

\noindent \textbf{Dynamic Object Reconstruction with Novel-View Augmentation.} 
Although deformable 3DGS can provide frame-to-frame correspondences, it relies heavily on accurate multi-view observations of a dynamic scene for reconstruction. 
However, most video clips are typically captured from a certain view, lacking contents from other viewpoints across frames, which in turn leads to unsatisfactory reconstruction \cite{yang2024deformable,liu2024dynamic}. 
To address this, we propose to leverage multi-view diffusion models \cite{voleti2025sv3d} to assist in dynamic object reconstruction.
Given a video $X^{src} = \{ X_{i}^{src} | i =1 \text{,} 2 \text{,} \dots \text{,} T\}$, where T is the number of frames, we utilize a multi-view generator $\mathcal{M}_{\text{diff}}$ to generate $N$ (imperfect) novel views $\mathcal{V} = \{ V_{i,j} \mid i = 1, 2, \dots, T; \, j = 1, 2, \dots, N \}$ with associated camera pose. We can then combine these novel views with input frames to optimize the reconstruction of the target object.

\noindent \textbf{Balanced View Sampling.} 
Since 3D serves merely as a proxy during the video editing process, users primarily focus on editing the observed view $X^{\text{src}}$, while the generated novel views  $\mathcal{V}$ are employed mainly to enhance the completeness of the reconstruction. 
However, trivially combining them inevitably amplifies inter-frame and intra-frame inconsistencies arising from the target object variations and generated novel views, leading to incomplete or irregular geometry (See row-2 in \figref{fig:ablation_4d}). 
To this end, we introduce a Balanced View Sampling to alleviate the inconsistency between observed and generated views.
We define two sampling probabilities: $\beta_i$ for each observed view $X^{src}_i$, and $\zeta_{i\text{,}{j}}$ for each novel view $V_{i\text{,} j}$ in $\mathcal{V}$, representing their respective selection probabilities during sampling.
Since the number of novel views is $N$ times larger than that of the observed view at each frame $i$, an equal sampling strategy would disproportionately favor the novel views, amplifying inconsistencies due to their synthetic nature.
Thus, we enforce the constraint that the total sampling probability of all views in $\mathcal{V}_i$  equals to  that of the observed view $X^{src}_i$, \ie, $\beta_{i} \;=\; \sum_{j=1}^{N}\zeta_{i\text{,}j}$.
This inherently assigns a lower sampling frequency to each novel view in $V_{i}$ compared to $X^{src}_i$, reducing inconsistencies caused by the novel views.

\noindent \textbf{Loss function.} 
The overall training objective is:
\begin{equation}
    \mathcal{L}=\mathcal{L}_{gs} + \mathcal{L}_{mask} + \mathcal{L}_{rgb} + \mathcal{L}_{depth} \text{.}
\end{equation} 
We employ a combination (\ie, $\mathcal{L}_{gs}$) of L1 loss and SSIM loss on the Gaussian images, 
using the input video frames and generated novel views as supervision.
To constrain the shape of the reconstructed mesh from different views, we apply an L1 loss (\ie, $\mathcal{L}_{mask}$). Additionally, we adopt the same rendering loss as $\mathcal{L}_{gs}$ to supervise the mesh image (\ie, $\mathcal{L}_{rgb}$). 
To mitigate sunken surfaces and floating artifacts in reconstructed meshes,
we further enforce a scale-invariant depth constraint $\mathcal{L}_{depth}$ between mesh depth and GT image depth.

\subsection{Editing 3D Proxy with Automatic Propagation}
\label{subsec:editing}
To enable a user-friendly editing process,  we aim to transfer cross-frame correspondences in deformable Gaussian splatting to the reconstructed mesh, such that users can edit the canonical mesh once, with the edits propagated to all frames in 3D space automatically. 
Formally, given a canonical Gaussian $G_c$, we first convert it to the canonical mesh $M_c$ via Marching Cube (MC). 
The user then performs edits directly on $M_c$, yielding the edited canonical mesh $M_c^{'}$ and thus the editing offset in the canonical space is $\Delta_m = M_c^{'} - M_c$.

\begin{figure} 
 \centering
\includegraphics[width=0.5\textwidth]{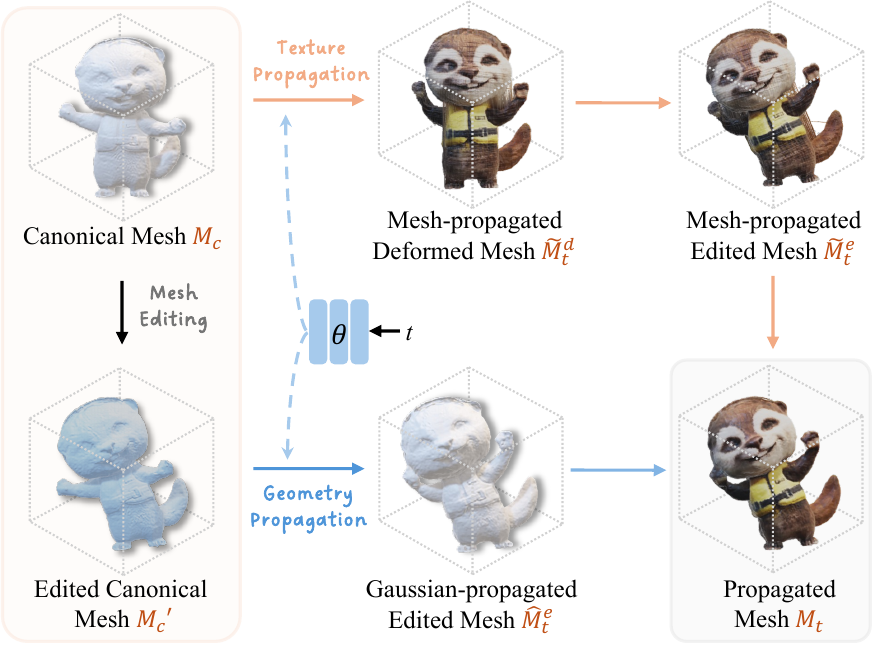}
    \caption{Workflow of our consistent editing propagation. Solid orange arrows denote texture propagation from the canonical mesh, and solid blue arrows indicate geometry propagation from canonical Gaussians.}
  \label{fig:stage2}
\end{figure}

A naive approach to propagate this editing offset is to directly apply the deformation field \footnote{For brevity, we omit the $t$ symbol in time-dependent MLPs, \eg, deformation field $\theta(\cdot)$ and color MLP network $MLP_c(\cdot)$.} $\theta(\cdot)$ to the $M_c$ to obtain the mesh-propagated deformed mesh $\Tilde{M}_t^{d} = M_c+\theta(M_c)$ at each frame $t$, and then integrate the $\Tilde{M}_t^{d}$ with the editing offset to produce the mesh-propagated edited mesh $\Tilde{M}_t^{e}=\Tilde{M}_t^{d} + \Delta_m$ (see flow indicated by the orange arrow in \figref{fig:stage2}). 
However, since the deformation field is optimized for Gaussian points rather than mesh vertices directly, certain vertices may incur positional shifts, leading to inaccurate geometry (see \figref{fig:dsq}, Variant-1). 
To this end, we develop a Dual Propagation Strategy that propagates geometry and texture edits using the canonical Gaussians and the canonical mesh, respectively.

\noindent \textbf{Geometry Propagation.} 
As the user's editing is performed in the canonical mesh,  we first build a vertex-point mapping to transfer the offset $\Delta_m$ for the canonical mesh to $\Delta_g$ for the canonical Gaussian.

We introduce a nearest-neighbor mapping from $G_c$ to $M_c$. For any gaussian point $\mathbf{x}$ in $G_c$, we  seek the closest vertex in $M_c$ by:
\begin{equation}
    \phi(\mathbf{x})
    \;=\;
    \arg\min_{\mathbf{v} \,\in\, M_c}
    \|\mathbf{x} - \mathbf{v}\|_2.
    \label{eq:mapping1}
\end{equation}
Thus, the editing offset for canonical Gaussian is $\Delta_g = \Delta_m(\phi(G_c))$.
Then, we leverage the correspondence in the deformation field $\theta(\cdot)$, to obtain the deformed Gaussian $G_t = G_c + \theta(G_c)$. 
The Gaussian-propagated edited mesh at frame $t$ can thus be produced by integrating $G_t$ with the editing offset $\Delta_g$ by $\hat{M}_{t}^{e}=\text{MC}(G_t + \Delta_g)$.

\noindent \textbf{Texture Propagation.} 
With $\hat{M}_t^{e}$, one naive method of obtaining the edited texture is to query the color directly by using it as input to the color network $MLP_{c}(\cdot)$. However, due to vertex misalignment before and after the editing, this method often results in spatially shifted colors (see \figref{fig:dsq}, Variant-2).

Instead, we note that although the mesh-propagated edited mesh $\Tilde{M}_t^{e}$, may contain geometric errors due to outliers in vertices and faces, the majority of its color remains correct.
Thus, we establish an additional mapping $\xi(\cdot)$ (similar to \reqref{eq:mapping1}) between the two edited meshes, $\Tilde{M}_t^{e}$ and $\hat{M}_t^{e}$, which are propagated from the canonical mesh and canonical Gaussians, respectively. We can then retrieve the color information from $\Tilde{M}_t^{e}$ and map it to $\hat{M}_t^{e}$.
Specifically, we first obtain the color $\Tilde{C}_t^{d}$ of deformed mesh $\Tilde{M}_t^{d}$ before editing. 
As the number of vertex is the same for $\Tilde{M}_t^{d}$ and $\Tilde{M}_t^{e}$ before and after geometry editing, the color of the mesh-propagated edited mesh can be queried by:
\begin{equation}
    \Tilde{C}_t^{e} =\Tilde{C}_t^{d} =MLP_{c}\bigl(\Tilde{M}_t^{d} - \theta^{-1}(\Tilde{M}_t^{d})\bigr),
    \label{eq:texture1}
\end{equation}
where $\theta^{-1}(\cdot)$ is the backward deformation field. 
Then, we can obtain the propagated texture color  $\hat{C}_t^{e} \;=\; \Tilde{C}_{t}^{e}\bigl(\xi(\hat{M}_t^{e})\bigr)$ for the Gaussian-propagated edited mesh $\hat{M}_t^{e}$, and form the propagated mesh $M_t=\{\hat{M}_t^{e};\hat{C}_t^{e}\}$.
%
Finally, we render it into its 2D representations: normal map for geometry and mesh image for texture.

\subsection{Generative Rendering from the 3D Proxy}
\label{subsec:render}

\begin{figure}[t!]
\centering
  \includegraphics[width=0.47\textwidth]{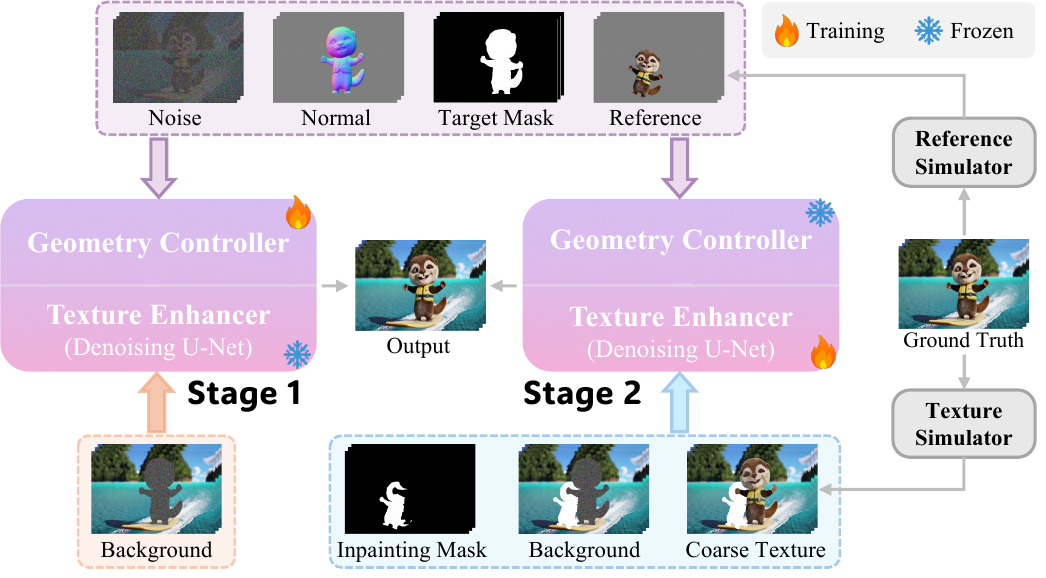}
  \caption{Training pipeline of the generative rendering from the 3D proxy. Using rendered geometry (\ie, normal map) and texture as separate control signals, we adopt a self-supervised mixed training strategy in which the geometry controller and the texture enhancer (\ie, denoising U-Net) are alternately trained in two stages. The contents enclosed by the purple dashed box indicates inputs shared by both stages, while those in the orange and blue dashed boxes are specific to the first and second stages, respectively. Reference objects and coarse textures are constructed via the reference and texture simulators to facilitate self-supervised training.
  }
  \label{fig:stage3}
\end{figure}

With the edited geometry and texture in place, the next step is to synthesize the final refined video renderings by leveraging these signals along with the input frames. 
To this end, we train a decoupled video diffusion model using a self-supervised mixed training strategy. The overall training pipeline is illustrated in \figref{fig:stage3}.

\begin{figure*}[t!]
  \includegraphics[width=0.95\textwidth]{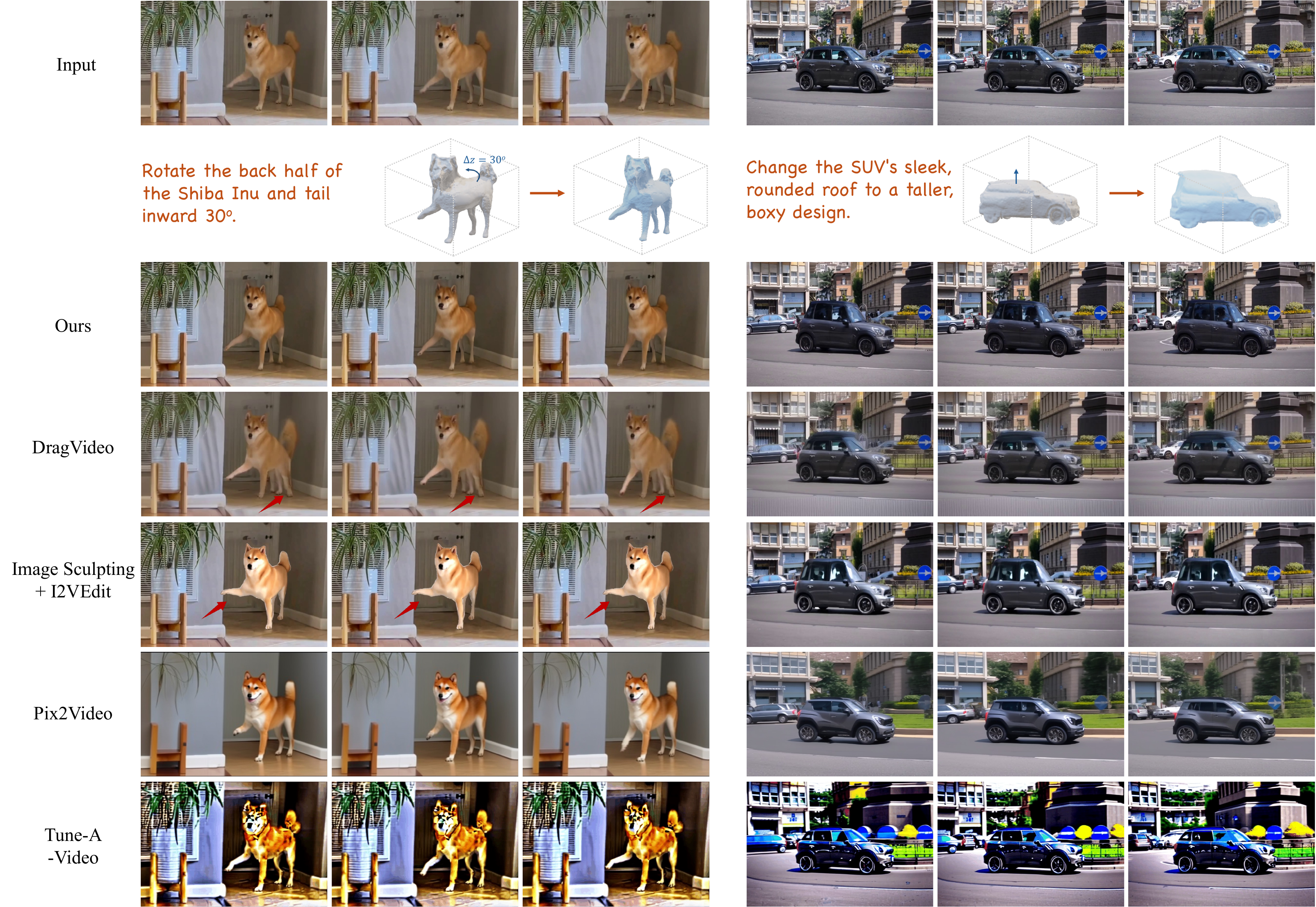}
  \caption{Qualitative comparison with four video editing baselines (DragVideo~\cite{deng2025dragvideo}, Image Sculpting~\cite{yenphraphai2024image} + I2V-Edit~\cite{ouyang2024i2vedit}, Pix2Video~\cite{guo2024i2v}, and Tune-A-Video~\cite{wu2023tune}) on our \dataset dataset samples. Due to their lack of 3D awareness, the baseline methods only achieve limited modifications to the object geometry. Instead, our approach enables direct editing in the 3D space only once, ensuring precise and consistent results for all frames. The first row shows inputs, and the second highlights text instructions (provided solely for illustrating the performed edits) and 3D space edits. 
}
  \label{fig:compare}
\end{figure*}

\noindent \textbf{Data Construction.} 
Since no such 3D-video paired video editing datasets are available, we generate training data by augmenting the input video to simulate pre- and post-editing states. 
(1) \textit{Pre-editing:} The reference object is constructed by the \textit{reference generator}, which first segments the target object from the original video and then applies random augmentations, including scaling, shifting, and rotation. 
(2) \textit{Post-editing:} The original video frames serve as Ground Truth (GT), with masked normal maps extracted as geometry control. Then, the \textit{texture simulator} generates coarse texture by randomly degrading the GT with SLIC-based segmentation~\cite{achanta2012slic}, median blurring, and down-up sampling.

\noindent \textbf{Overall Pipeline.} 
We use the I2V release of the stable video diffusion (SVD) \cite{blattmann2023stable} as the base model, and follow the ControlNet~\cite{zhang2023adding}  paradigm. 
To handle the domain gap between texture and geometry, we employ the base model to refine coarse texture, and utilize the control branch to guide the structure via geometry control. Refer to Supp. C.2 for more details.

\noindent \textbf{Mixed Training.} 
A straightforward approach is to train the entire model jointly using two control signals. 
Nonetheless, the coarse texture is pixel-aligned to the GT, while the distribution of the geometry is very different. It is easy to discard the geometry branch and downgrade the generative model to an enhancement model again. 
Motivated by the animation production that first draws structure and then coloring, we propose to train the geometry controller and texture enhancer (\ie, base model) in a two-stage manner. 

Specifically, in the first stage, we fix the base model and train only the geometry controller. In the second stage, we freeze the trained geometry controller and fine-tune the base model. 
In both stages, the model receives the same set of shared inputs: random noise, a normal map for geometry control, a target mask, and a reference object~\footnote{For simplicity, we omit the process of transforming pixels to latents through the VAE.}. 
The primary difference between the two stages lies in the handling of texture control. In the first stage, texture control is replaced by a background image, where random noise is inserted into the object region.
In this way, although the generated appearance may appear coarse (see \figref{fig:ablation_2}, Case-2) due to the spatial misalignment between the reference object and the target geometry, the model is encouraged to learn geometry-following generation.

In the second stage, the model focuses on texture enhancement by simultaneously refining details within the edited object and inpainting missing regions (\eg, areas filled with white in \figref{fig:stage3}). 
Unlike the first stage, this phase introduces an additional set of stage-specific inputs, \ie, an inpainting mask and a coarse texture rendered from the edited 3D proxy. 
To preserve the model’s ability to follow geometric guidance, the texture control is provided as input with a probability of 20\%, while it is replaced by a background image with a probability of 80\%. 
This mixed-training strategy encourages the model to balance appearance preservation when texture information is available, while retaining generative flexibility when it is not.

\section{Experiments}
\label{sec:exp}

\begin{table}[t!]
\centering
\caption{Quantitative comparison of video editing methods.  The user study reports the average rank in editing quality (EQ), semantic consistency (SC), and visual plausibility (VP). }
\vspace{-2mm}
\setlength{\tabcolsep}{3pt} 
\scalebox{0.85}{ 
\begin{tabular}{l|ccc|ccc}
\toprule
\multirow{2}{*}{Methods} & \multicolumn{3}{c|}{ Metrics } & \multicolumn{3}{c}{User Study } \\
 & Fram-Acc $\uparrow$ & Tem-Con $\uparrow$  & CLAP Score $\uparrow$  & EQ $\downarrow$ & SC $\downarrow$& AP $\downarrow$ \\
\midrule
Tune-A-Video  &  0.510 &	0.979&	0.423&	5.35&	5.68&	5.85  \\
Pix2Video  &  0.825 &	0.988 &	0.647&	4.81&	5.03&	4.65    \\
I2V-Edit  &0.856 &	0.989&	0.792&	2.23&	2.33&	2.23   \\
DragVideo  &  0.875&	0.983&	0.774&	3.27&	2.72&	2.68    \\
VideoShop & 0.812 & 0.985 & 0.726 & 4.21 & 3.88 & 3.78 \\
Ours &   \textbf{0.945}& \textbf{0.990}& \textbf{0.878}& \textbf{1.13}& \textbf{1.36}& \textbf{1.81} \\
\bottomrule
\end{tabular}
}
\label{tab:sota}
\vspace{-2mm}

\end{table}

\subsection{Comparisons with State of the Art Methods}
\label{subsec:sota_comp}
\subsubsection{Qualitative Results.} 

To the best of our knowledge, ours is the first work to leverage a 3D proxy for video editing. As existing methods lack such precise editing capabilities, we compare our approach against four baseline methods (\ie, 
Tune-A-Video~\cite{wu2023tune},
Pix2Video~\cite{guo2024i2v},  
Image-Sculting+I2V-Edit~\cite{ouyang2024i2vedit}), 
DragVideo~\cite{deng2025dragvideo}, and Videoshop~\cite{fan2024videoshop}+Zero-1-2-3~\cite{shi2023zero123++}.

\figref{fig:compare} shows the visual comparison. It clearly demonstrates that when dealing with fine-grained video editing scenarios, such as \textit{locally rotating the dog} in case-1 and \textit{stretching the car's roof} in case-2, all compared methods exhibit varying degrees of failure. 
Among the compared methods, I2V-Edit produces acceptable geometry editing results, due to reliable 3D editing on the first frame \footnote{For a fair comparison, we use the 3D editing-capable Image-Sculpting to edit the first frame.}.  
However, as it heavily relies on the first frame for propagation, it often fails when there is a constant geometry/shape variation across frames. For example, the dog's leg hangs across all frames in case-1. 
Moreover, since DragVideo only supports point-to-point editing in the 2D space, its limited point coverage is insufficient for fine-grained local editing. As a result, it can extend the car’s roof but fails to account for the entire upper part of the car in case-2, and generates multiple legs for the dog in case-1.  
As for the prompt-based methods, neither the training-free Pix2Video nor the optimization-based Tune-a-Video can handle fine-grained editing with text prompts. 
In contrast, our approach achieves high-quality  results due to precise and consistent editing in the 3D space and reliable enhancement in the 2D space.

\subsubsection{Quantitative Results.}

We collect \dataset to evaluate our method. \dataset consists of 70 videos, covering six categories: pose editing, rotation, scaling, translation, texture modification, and object composition. 
For evaluation, we employ widely used CLIP~\cite{radford2021learning}-based metrics, Fram-Acc~\cite{ma2023magicstick} and Tem-Con~\cite{esser2023structure}. 
However, for fine-grained edit types, CLIP cannot measure the appearance consistency of objects before and after editing. 
To this end, we also introduce CLAP (CLIP-APpearance) Score, to measure the cumulative error of the Fram-Acc and the DINO similarity accuracy: $\text{CLAP Score} = \text{Fram-Acc} \times \text{DINO}(input, output)$, where $input$ and $output$ represent the input and edited frames.
\tableref{tab:sota} shows that our method achieves the highest performance, demonstrating the superior editing quality and temporal consistency compared to the baselines. 
We also conduct a user study to evaluate the perceptual quality in editing quality (EQ), semantic consistency (SC), and visual plausibility (VP). The results indicate that our method consistently outperforms others and is the most preferred.

\begin{table}
\caption{Ablation study on consistent 3D proxy reconstruction. 
All ablations are trained and tested on 21 novel views. \textit{BVS:} balanced view sampling.}
\vspace{-2mm}
    \centering
    \vspace{0pt}
    \scalebox{0.9}
    {
    \begin{tabular}{ccc|cccc}
    \toprule
   \textit{Novel Views} & \textit{BVS} & $L_{d}$& PSNR $\uparrow$ & SSIM $\uparrow $ & LPIPS $\downarrow$ & DINO Score $\uparrow$ \\
    \midrule 
     & & & 20.35 & 0.903 & 0.060 & 0.375 \\
    \checkmark &  & & 22.87 & 0.922 & 0.057 & 0.396 \\
    \checkmark & \checkmark &  & 22.91  & 0.923 & 0.055 & 0.396 \\
   \checkmark & \checkmark & \checkmark & 23.00 & 0.923 & 0.055 & 0.397 \\
    \bottomrule
    \end{tabular}
    }
    \label{tab:abl_4d}
    \vspace{-4mm}
\end{table}

\begin{table}[t!]
\centering
\vspace{2mm}
\caption{Ablation study of decoupled video diffusion.}
\vspace{-2mm}
\setlength{\tabcolsep}{4pt} 
\scalebox{0.95}{ 
\begin{tabular}{cc|ccc}
\toprule

 Stage-1 & Stage-2 & Fram-Acc $\uparrow$ & Temp-Con $\uparrow$  & CLAP Score  $\uparrow$ \\
\midrule
 \checkmark & &  0.758 &  0.987   & 0.723   \\
 & \checkmark &  0.873 &  0.988   & 0.769   \\
\checkmark  & \checkmark   &  0.945 &  0.990  & 0.878  \\

\bottomrule
\end{tabular}
}
\label{tab:ablation_svd}
\vspace{-3mm}
\end{table}

\begin{figure*}[t!]
  \includegraphics[width=0.97\textwidth]{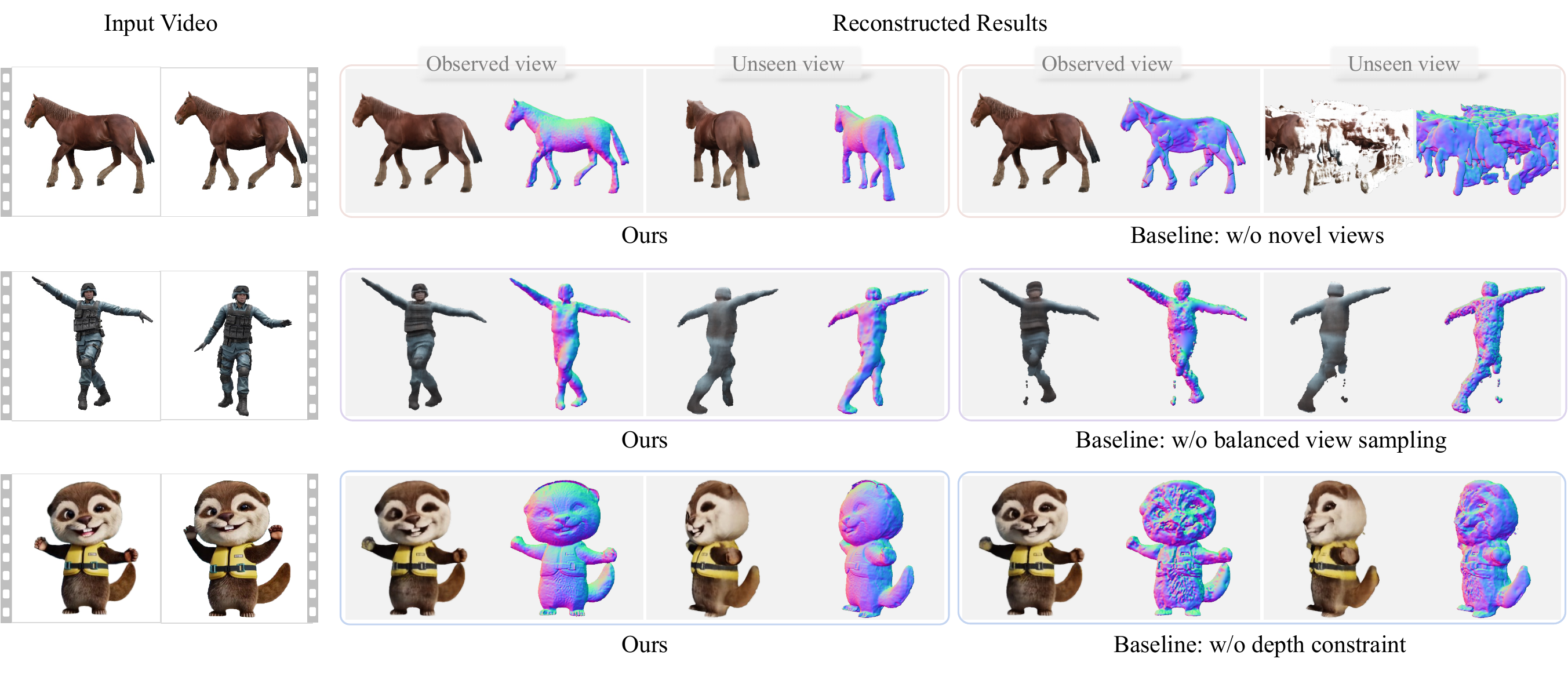}
  \vspace{-4mm}
  \caption{Comparison of the key components in consistent 3D proxy  reconstruction. Each row starts with an input video on the left, followed by results of our full model in the middle and ablated baselines on the right, each consisting of a rendered image and its corresponding normal map. }
  \label{fig:ablation_4d}
\end{figure*}

\begin{figure*}
  \includegraphics[width=0.96\textwidth]{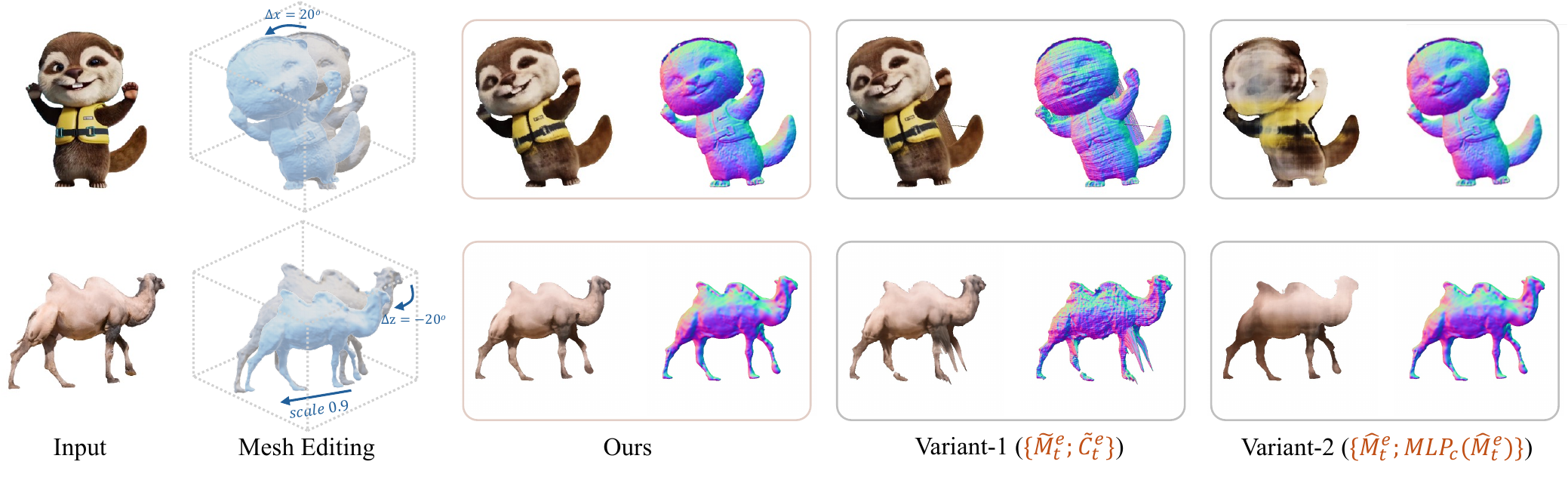}
  \vspace{-3mm}
    \caption{Comparison of editing propagation strategies. The edited 3D mesh is rendered into the texture (\ie, mesh image) and geometry (\ie, normal map). The first column presents the input image at frame $t$, followed by the editing in 3D space at this frame in the second column, with results from ours and two variants shown in subsequent columns. 
    Variant-1: mesh-propagated edited mesh with its color. Variant-2: Gaussian-propagated edited mesh with its color.
    }
  \label{fig:dsq}
\end{figure*}

\begin{figure*}[ht!]
  \includegraphics[width=0.95\textwidth]{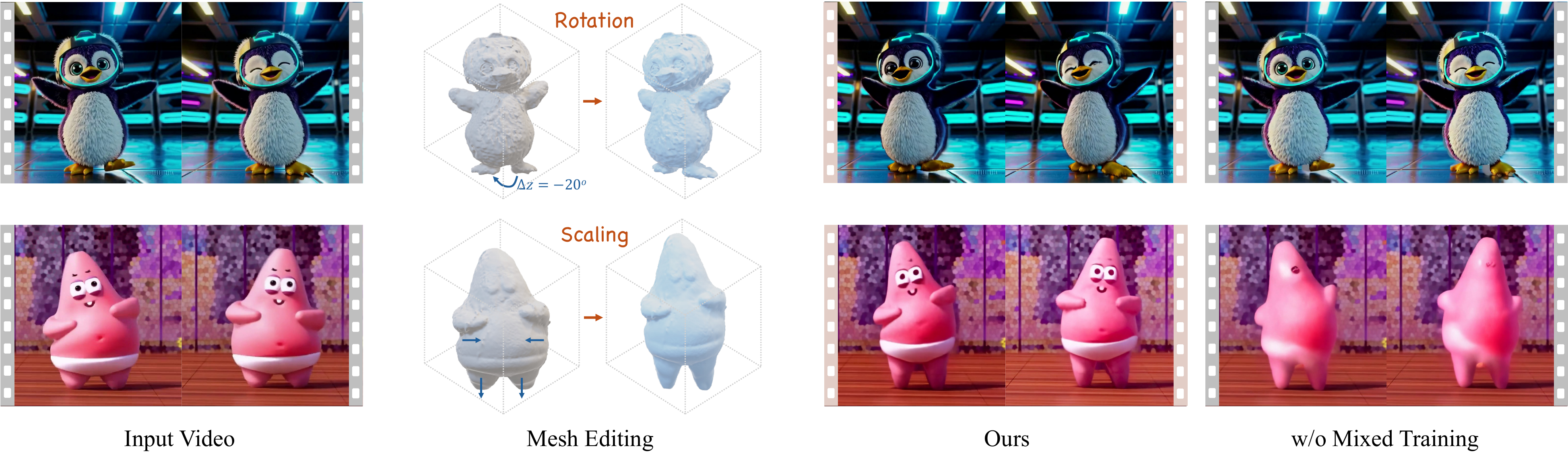}
  \vspace{-3mm}
    \caption{Comparison of mixed training in generative rendering from the 3D proxy. For each case, the input video is shown on the left, followed by 3D space editing in the second column, with results from our method and the baseline in the subsequent columns. Two frames are displayed for each case.}
  \label{fig:ablation_2}
\end{figure*}

\subsection{Ablation Study}
\subsubsection{Consistent 3D Proxy Reconstruction}
We validate each component by individually removing them, with visual results shown in \figref{fig:ablation_4d}. It shows that the absence of novel views as constraints leads to wrong reconstructions (row 1). 
Disabling the balanced sampling strategy increases the sampling frequencies for novel views, which, in turn, results in greater inconsistencies and incomplete reconstructions (row 2). 
When the depth loss is removed, the mesh images remain relatively unaffected due to their depth-invariance, but the normal map exhibits numerous sunken holes (row 3).

The results are presented in \tableref{tab:abl_4d} in terms of PSNR, SSIM, LPIPS, and DINO score. 
We can conclude that (1) Using only the monocular video results in poor reconstruction quality, while introducing novel views significantly improves all metrics (row 1 \textit{vs.} row 2), highlighting the critical role of novel views as constraints; (2) Further incorporating balanced view sampling and depth constraints leads to additional performance gains (row 3 \& 4).

\subsubsection{Editing 3D Proxy with Automatic Propagation}

We compare two variants for propagating edits. 
(1) We directly utilize the mesh-propagated edited mesh $\Tilde{M}_t^{e}$  combined with its vertex color $\Tilde{C}_t^{e}$  as the propagated 3D mesh $\{\Tilde{M}_t^{e}; \Tilde{C}_t^{e}\}$ at frame $t$. 
(2) After obtaining the Gaussian-propagated edited mesh $\hat{M}_t^{e}$, we send it to the color network $MLP_c(\cdot)$ to query its vertex colors and get $\{\hat{M}_t^{e}; MLP_c(\hat{M}_t^{e})\}$. 

We present the comparisons of geometry and texture renderings in \figref{fig:dsq}.
Although the deformation field originally designed for GS points can be applied to mesh vertices, discrepancies in quantity and position between GS points and mesh vertices lead to several inaccuracies in the reconstructed geometry and texture. 
In contrast, the second variant naively queries colors, leading to noticeable color shifts. Our method, instead, establishes a dual propagation for both geometry and texture, thus achieving higher editing consistency.

\subsubsection{Generative Rendering from 3D Proxy.} 
We evaluate the effectiveness of the mixed training strategy by disabling Stage 2 and present the visual comparisons in \figref{fig:ablation_2}. 
Training exclusively on augmented data causes the model to struggle with view-changing edits (\eg, rotating the penguin 20 degrees to the left) when relying solely on geometry as control. This is because such paired data cannot be accurately simulated during the augmentation process.
Additionally, spatial misalignment between the target geometry and the reference object leads to failures in maintaining the correct appearance (\eg, Patrick Star’s face and pants are distorted or lost).
Quantitative comparisons in \tableref{tab:ablation_svd} on the \dataset dataset also demonstrate that our mixed training strategy significantly improves the Frame-Acc and CLAP Score metrics.

\subsection{Discussion}
\label{subsec:discussion}

\begin{figure}[t!]
  \includegraphics[width=0.47\textwidth]{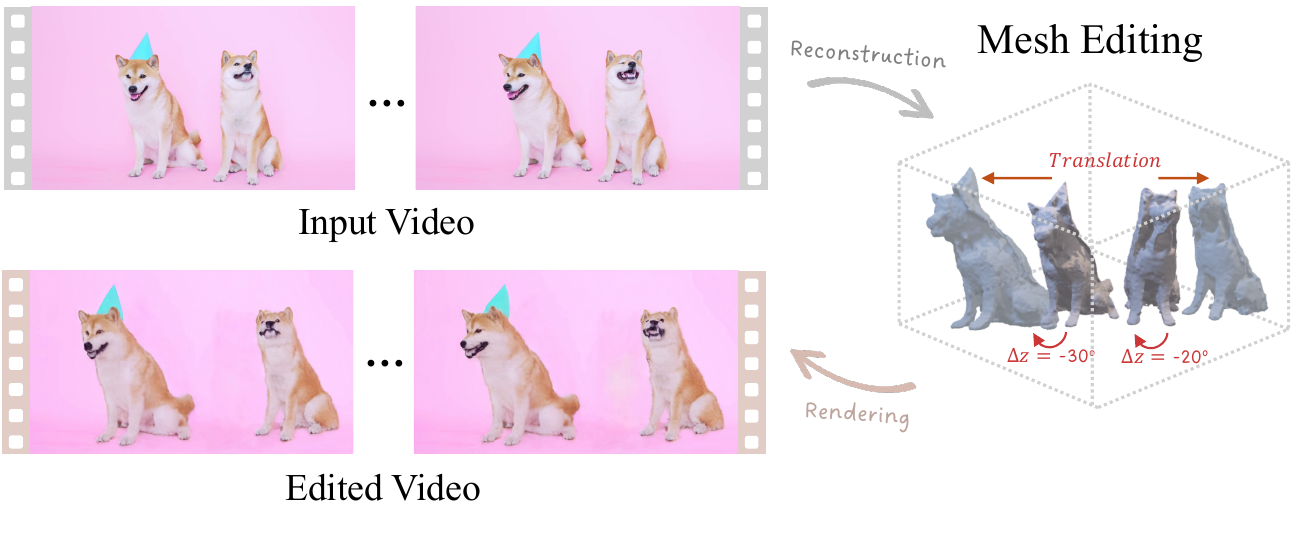}
  \vspace{-3mm}
  \caption{Visual illustration of multi-objects editing.
  }
  \label{fig:multi_objects}
  
\end{figure}

\subsubsection{Efficiency.} 

We report the runtime and inference cost. The reconstruction stage takes approximately 90 minutes on a single A100 GPU, while generative rendering requires 7 days of training on 8 A100 GPUs. At inference, for 14 frames of 512×512 resolution video, our mesh editing takes between 30 seconds and 10 minutes, depending on the task complexity, and generative rendering takes around 43 seconds on a single A100 GPU. In terms of GPU memory usage, reconstruction requires 12GB, while diffusion training and inference consume 70GB and 22GB, respectively.

\subsubsection{Multi-Object Editing.}

Our method focuses on addressing single-object editing but is adaptable to multi-objects. The reconstruction phase leverages a multi-view generator~\cite{voleti2025sv3d}, optimized for single-object contexts, necessitating individual reconstruction for each object. 
Objects are then manually edited, with edits propagated via offsets, and finally processed sequentially by our decoupled video diffusion model. \figref{fig:multi_objects} shows an example.

\begin{figure}[t!]
  \includegraphics[width=0.47\textwidth]{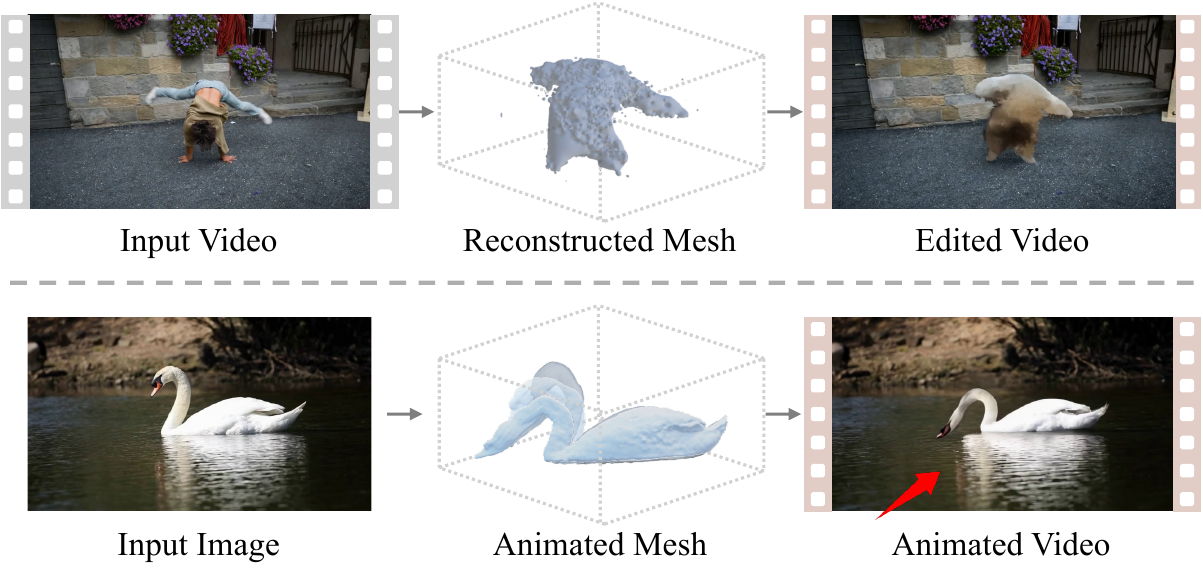}
  \vspace{-3mm}
  \caption{The top row illustrates failure cases where our framework struggles to reconstruct and edit objects with highly complex motion and invisible parts. The bottom row shows limitations in handling shadows and reflections.}
  \label{fig:limitation}
  \vspace{-1mm}
\end{figure}

\section{APPLICATIONS}
\label{sec:application}

\begin{figure*}
  \includegraphics[width=1\textwidth]{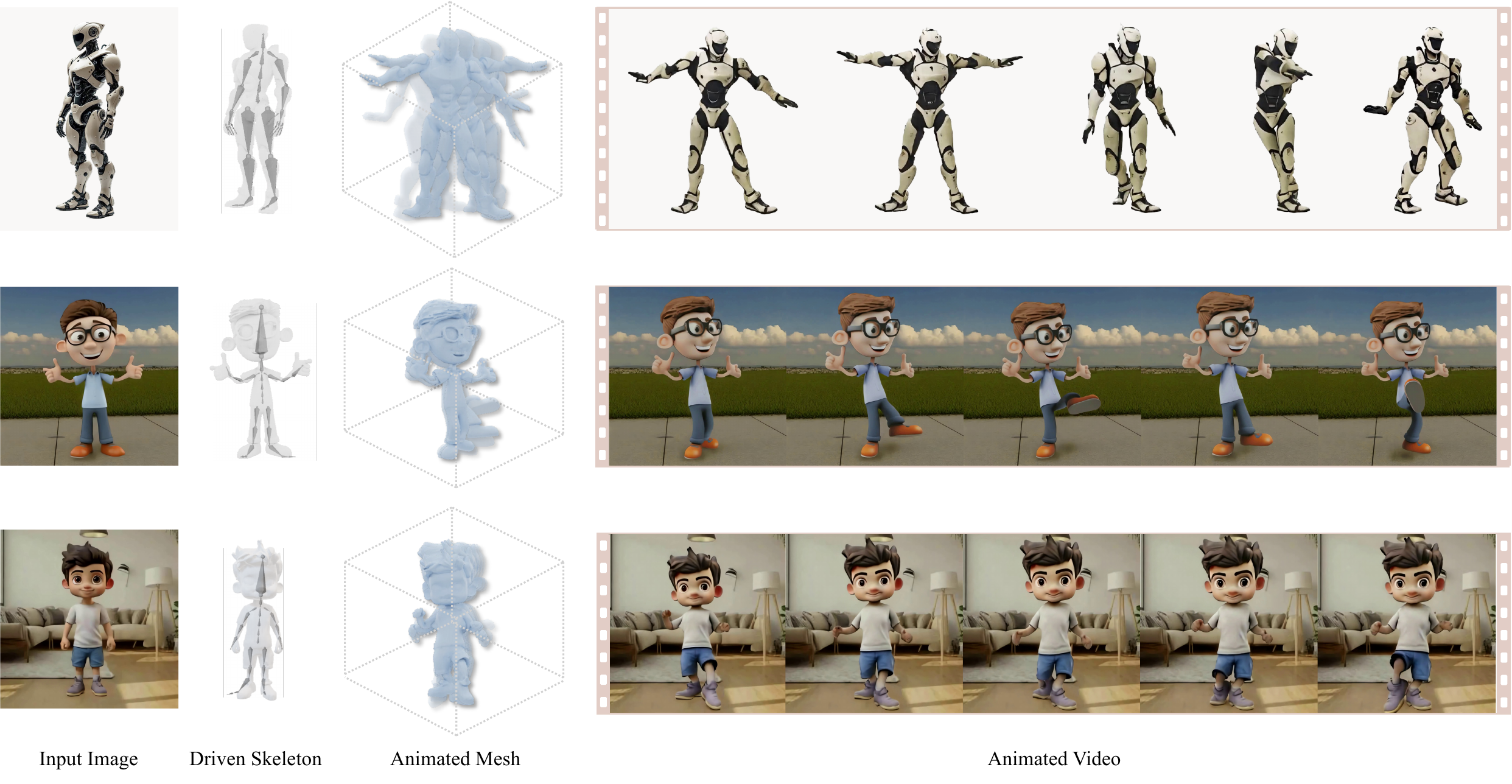}
  \vspace{-7mm}
    \caption{Our approach enables Image-to-Video animation: from a single input image (col. 1), we reconstruct a 3D mesh, rig it with a skeleton (col. 2), and animate it using motion sequences to produce edited meshes (col. 3). Final videos are rendered from the animated geometry and textures.
  }
  \vspace{-1mm}
  \label{fig:app_1}
\end{figure*}

\noindent \textbf{Image-to-Video Animation.} 
Given an input image, we first reconstruct a 3D model of the target object. The reconstructed 3D model can then be rigged and animated to create various motions. Finally, the edited 3D meshes are further employed by our decoupled video diffusion model to generate the animated video (see \figref{fig:app_1}).

\begin{figure*}
\includegraphics[width=1\textwidth]{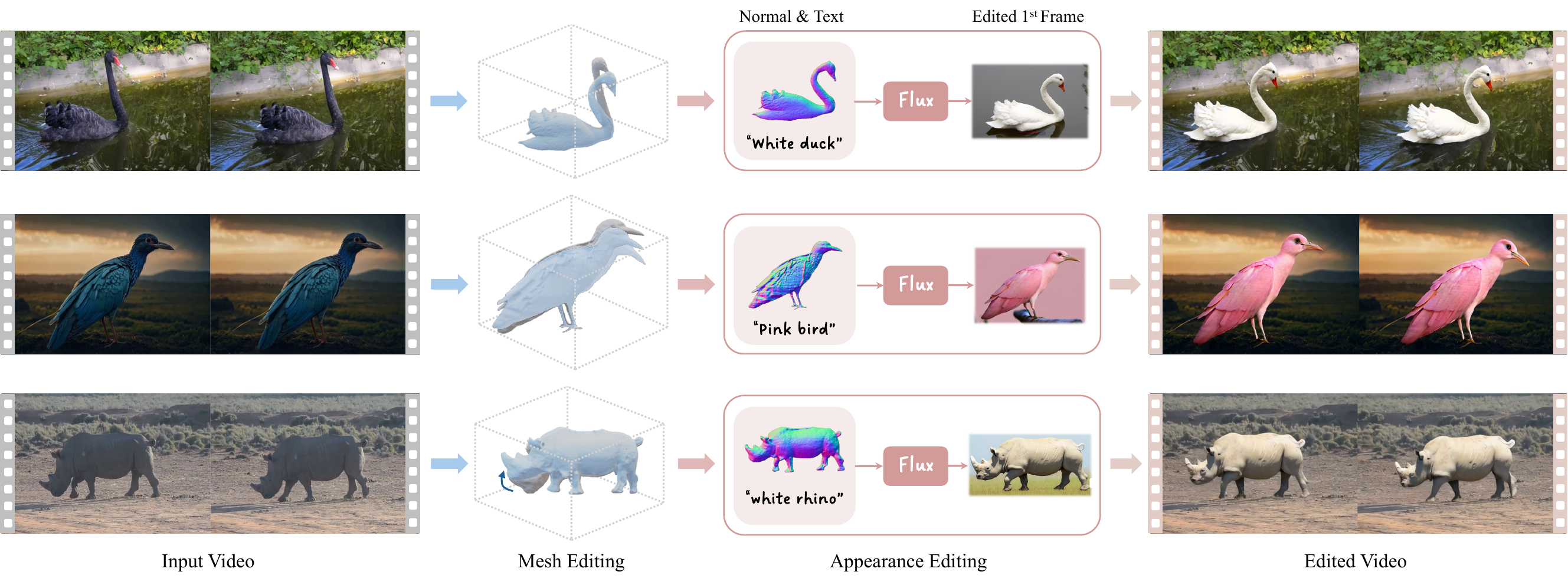}
  \vspace{-7mm}
  \caption{Our method integrates with existing 2D editing tools (e.g., Flux~\cite{flux2024}) for flexible appearance editing. Left: input frames; center: 3D geometry + 2D appearance edits; right: final results.}
  \vspace{-3mm}
  \label{fig:app_2}
\end{figure*}

\noindent \textbf{Appearance Editing.} 
As the geometry and texture controls are decoupled in our stage 3, we can easily integrate various image editing tools into our framework to support flexible object appearance editing. 
Visual results are presented in \figref{fig:app_2}.

\section{CONCLUSION}
In this work, we introduce \name, a 3D-aware video editing method that enables users to perform precise and consistent edits on target objects in a video. Our method leverages generated novel views as constraints to reconstruct the target object into a time-consistent mesh, allowing users to perform edits directly in 3D space with a dual-propagation strategy. 
With the help of 3D proxy, the proposed method demonstrates compelling performance in various video editing applications. 

\textbf{Limitation.} Our approach does have limitations. First, as illustrated in \figref{fig:limitation}, it encounters difficulties during the reconstruction stage when dealing with objects exhibiting complex motions. Second, it struggles to handle object associations. In the future, we aim to introduce the Visual Association Model~\cite{liu2024diff} and the Large Reconstruction Model to solve these problems.

\appendix

\renewcommand\thefigure{\Alph{section}\arabic{figure}}  
\renewcommand\thetable{\Alph{section}\arabic{table}}

\vspace{10mm}
\noindent This supplemental material provides additional details on our methods and experiments. \textbf{\secref{sec:supp_step1}} covers 3D proxy reconstruction, including data preparation and implementation. \textbf{\secref{sec:supp_step2}} focuses on editing 3D proxies with automatic propagation. \textbf{\secref{sec:supp_step3}} discusses generative rendering from 3D proxies, including inference workflow and experiments. \textbf{\secref{sec:supp_discussions}} includes a discussion on pipeline design, efficiency, mesh topology, reconstruction performance, and 3D proxy quality. Finally, \textbf{\secref{sec:supp_visual}} presents additional visual results.

\section{\textbf{Consistent 3D Proxy Reconstruction}}
\label{sec:supp_step1}

\subsection{Preliminaries} 
3D Gaussian Splatting (GS) \cite{kerbl20233d} adopts a novel approach based on explicit point clouds to efficiently model 3D scenes. Each 3D Gaussian is parameterized by its mean position $\mu$, a covariance matrix $\Sigma$ and other attributes. Formally, a 3D Gaussian $G$ is defined as: 
\begin{equation}
    G(x) = e^{-\frac{1}{2}(x)^T\Sigma^{-1}(x)}
\end{equation}
Each Gaussian is multiplied by opacity $\alpha$ during the blending process. When projecting 3D Gaussian to 2D space, the covariance matrix is updated to $\Sigma'$ using a Jacobian matrix $J$ and a viewing transformation $W$ via $\Sigma'=JW \Sigma W^{T}J^{T}$. 
To handle the differentiable optimization, the covariance matrix $\Sigma$ is divided into two learnable elements $r$ and $s$ to represent the rotation and scaling, which is then transformed into the corresponding matrices $S$ and $R$ to form the $\Sigma$ via $\Sigma =RSS^{T}R^{T}$. Each 3D Gaussian is represented as $G(x \text{;} \mu \text{,} r \text{,} s \text{,} \alpha)$.

Deformable 3DGS \cite{yang2024deformable} extend the 3D GS to dynamic scene by learning a set of canonical Gaussians along with a time-varying deformation field $\theta$ parametrized as an MLP. 
At each time $t$, the position $\gamma(x)$ of 3D Gaussians and time $\gamma(t)$ with positional encoding are used as input to the deformation MPL to obtain the offset $\theta(x)$, $\theta(r)$ and $\theta(s)$ of the dynamic 3D Gaussians in canonical space. 
$\gamma(\cdot)$ denotes the positional embedding function. 
The new 3D Gaussian at the deformed space can then be represented as $G'(x+\theta(x) \text{;} \mu \text{,} r+\theta(r) \text{,} s+\theta(s) \text{,} \alpha)$.

Directly performing editing on the Gaussian Splatting has been explored \cite{shin2024enhancing,sun2024splatter} recently. However, manipulating irregular 3D points is not user-friendly and only supports limited editing types.

\subsection{Data Preparation}
\begin{figure*}[th!]
  \includegraphics[width=1\textwidth]{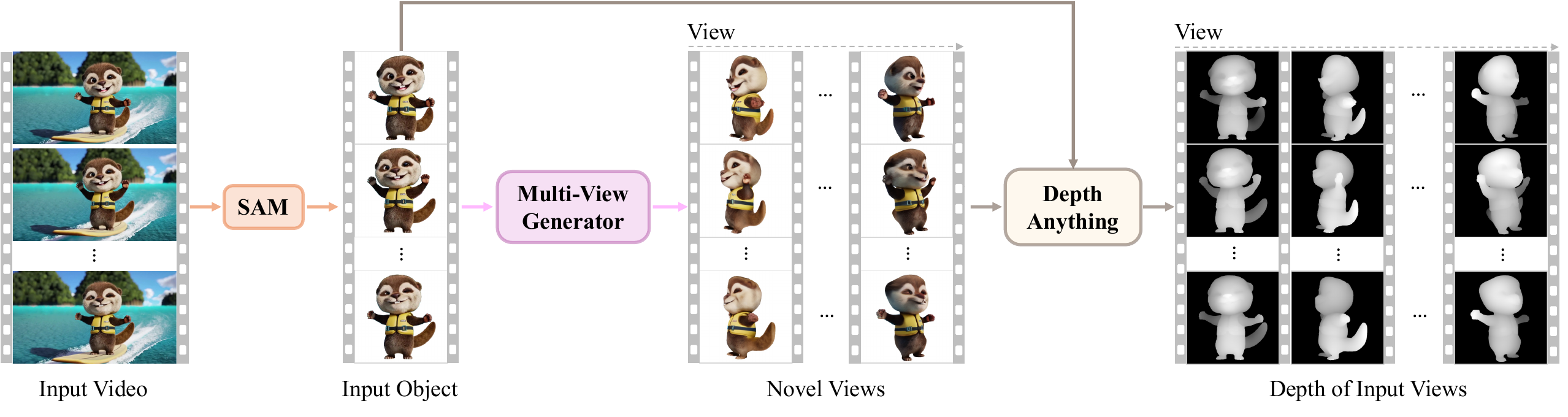}
  \caption{Workflow of data pre-processing in consistent 3D proxy reconstruction.}
  \label{fig:s1_workflow}
\end{figure*}

The workflow of data preparation during the target object reconstruction is depicted in \figref{fig:s1_workflow}. Given an input video, we first crop and segment the target object using SAM2~\cite{ravi2024sam}. Each frame of the cropped video is processed by a multi-view generator \cite{voleti2025sv3d} to produce novel views. 
During the novel views generation, we assume that the camera in the input video is fixed and set the same camera pose for all input frames, where the field of view (FOV) = 33.8, elevation angle = 0, and azimuth angle = 0. 
For the newly generated views, we keep the field-of-view and elevation angles the same as in the input frame and sample different azimuths from the 360-degree sphere. The six azimuths are: \{51.43, 102.86, 154.29, 205.71, 257.14, 308.57\}.
Then, we send both the original input video frames and the generated novel views to a depth estimation method \cite{yang2024depth} to produce depth maps, which will be used as the Ground-Truth depth supervision. 
We also employ SAM2 to obtain the Ground-Truth mask from the input frames and novel views. 
By default, we process 21 frames. If the motion between consecutive frames in the input video is minimal, the number of frames is increased to 42.

\subsection{Implementation Details}
\label{supp:gs_details}
Formally, given the input video frames $\mathbf{X^{\textit{src}}}$ and generated novel views $\mathbf{V}$, during the reconstruction process, we render a Gaussian-splatted image $I^{gs} \in \mathbf{R}^{H \times W \times 3}$ from the deformed gaussian, and render three mesh outputs from the Gaussian-propagated deformed mesh, \ie, mesh mask $ M^{pred}\in \mathbf{R}^{H \times W \times 1}$,  mesh depth $ D^{pred}\in \mathbf{R}^{H \times W \times 1}$ and mesh image $ I^{mesh}\in \mathbf{R}^{H \times W \times 3}$. The loss functions are illustrated as follows.

\noindent \textbf{GS Loss.} We follow \cite{kerbl20233d} to adopt a combination of L1 loss and SSIM loss to supervise the gaussian-splatted image:
\begin{equation}
    \mathcal{L}_{gs}=(1-\lambda_{ssim})\cdot||I^{gs}-I^{gt}||+\lambda_{ssim}\cdot\mathcal{L}_{ssim}(I^{gs}\text{,} I^{gt}) \text{,}
\end{equation}
where $\lambda_{ssim}=0.2$. We use the input frames and generated views as the $I^{gt}$ for the observed (input) and generated (novel) views, respectively.

\noindent \textbf{Mesh Mask Loss.} We apply an L1 loss to the rendered mesh mask to help constrain the shape of the mesh: 
\begin{equation}
    \mathcal{L}_{mask}=||M^{pred} - M^{gt}|| 
\end{equation}

\noindent \textbf{Mesh Image Loss.} We employ the same loss as $\mathcal{L}_{gs}$ to supervise the rendered mesh image $ I^{mesh}$:
\begin{equation}
    \mathcal{L}_{rgb}=(1-\lambda_{ssim})\cdot||I^{mesh}-I^{gt}||+\lambda_{ssim}\cdot\mathcal{L}_{ssim}(I^{mesh}\text{,}I^{gt}) \text{.}
\end{equation}

\noindent \textbf{Mesh Depth Loss.} Although the balanced view sampling strategy can alleviate inconsistencies and improve the overall quality of the 3D mesh, the surface geometry still suffers from depth ambiguities, resulting in issues such as sunken surfaces or floating artifacts. To address this, we further introduce a scale-invariant depth constraint:
\begin{equation}
    \mathcal{L}_{\mathrm{depth}}=\frac{1}{M^{gt}}\sum_{i=1}^M ||D_{i}^{\mathrm{pred}} - D_{i}^{\mathrm{gt}} || \text{,}
\end{equation}
where $M^{gt}$ denotes the Ground-Truth mask. $\mathbf{D}_{i}^{\mathrm{gt}}$ and  $\mathbf{D}_{i}^{\mathrm{pred}}$ are the disparity maps predicted by Depth-Anything \cite{yang2024depth} and  rendered from the reconstructed mesh  (\ie, $D^{pred}$), respectively.

During optimization, gradients can propagate back to the Gaussians and the deformation MLP through both the Gaussian splatted image and the mesh outputs, enabling updates to these components jointly. In this way, the correspondences across frames in the deformable GS are transferred to the mesh, resulting in a correspondence-enabled mesh for the target object.

\noindent \textbf{Implementation.} We use the implementation from 3D Gaussian Splatting \cite{kerbl20233d} for differentiable Gaussian rasterization. Instead of using  SfM for initialization, we initialize the 3DGS using points uniformly sampled from a sphere of radius 1. The model is trained for 25k iterations, with the first 3k optimizing only the 3D Gaussians for the first frame. Joint training of 3D Gaussians and the deformation field follows, and from iteration 12k, DPSR with Marching Cubes is introduced to optimize the mesh geometry.
By default, $N=6$ novel views are uniformly sampled along a circular trajectory using SV3D \cite{voleti2025sv3d}. All reconstructions were conducted against a white background at a resolution of $576\times576$ on an NVIDIA A100 GPU. 
For the optimized networks, such as the deformation field $\theta$ and the color network $MLP_c(\cdot)$, we follow the same implementations as in \cite{yang2024deformable}. Additionally, the backward deformation field $\theta^{-1}$ adopts the same network as $\theta$.   
Considering the inconsistency caused by the novel views, for all losses, we apply a smaller weight (1/5) to novel views than the observed view.

\noindent \textbf{Querying the color during training.}  
We adopt the approach of \cite{liu2024dynamic}, storing vertex colors in a canonical color MLP, denoted as $MLP_{c}(\cdot)$. The vertex color query process for the deformed mesh proceeds as follows:
\begin{itemize}
    \item \textbf{Deformation Query}: Given the canonical Gaussians $G_c$, we query their deformation of $G_c$ at time t via the deformation field $\theta(G_c)$.
    \item \textbf{Gaussian Wraping}: The deformed Gaussians  $G_t$ are computed as $G_t = G_c + \theta(G_c)$.
    \item \textbf{Mesh Extraction}: Using DPSR combined with Marching Cubes (MC), we convert $G_t$ into its corresponding deformed mesh.
    \item \textbf{Canonical Mapping}: The deformed mesh is projected back to the canonical space via the inverse deformation field $\theta^{-1}(\cdot)$.
    \item \textbf{Color Assignment}: Vertex colors are queried from $MLP_c(\cdot)$ in the canonical space and transferred to the deformed mesh, leveraging their one-to-one vertex correspondence.
\end{itemize}

\subsection{Additional Experiments}

We also conduct experiments to evaluate the use of novel views: (1) determining which time slots (\ie, frames) utilize novel views, with the number of novel views $N$ fixed; and (2) evaluating the impact of the number of novel views $N$, where novel views are applied to all frames. 
The results in \tableref{tab:abl_4d_novel_views} demonstrate that (1) providing novel views for only key frames is insufficient and (2) due to the inconsistencies introduced by novel views, using more novel views does not necessarily yield better results. Thus, by default, we utilize six generated novel views for each frame to balance the reconstruction completeness and novel-views inconsistency.

\begin{table}[t!]
\caption{Ablation study on the use of novel views. [f, m, l] represent the first, middle, and last frames, respectively. Note that all the metrics in this experiment are evaluated on six views. 
}
    \centering
    \vspace{0pt}
    \scalebox{0.9}
    {
    \begin{tabular}{l|cccc|ccc}
    \toprule
    \multirow{2}{*}{Metrics} & \multicolumn{4}{c|}{Time slots uses novel views} & \multicolumn{3}{c}{Number of novel views} \\
      & [f] & [f,m] & [f,m,l] &All &12 & 6 & 3 \\
    \midrule 
    PSNR $\uparrow$ & 22.53&    22.40& 22.60 &  23.45 & 23.59 & \textbf{24.04} & 23.30 \\
    SSIM $\uparrow$ & 0.919&     0.919& 0.921 & 0.925 & 0.926 & \textbf{0.928} & 0.925 \\
    LPIPS $\downarrow$ & 0.054&   0.054& 0.054 & 0.054 & 0.053 & \textbf{0.054} & 0.054 \\
    DINO Sim. $\uparrow$  &0.428&  0.445& 0.410 & 0.403 & 0.405  & \textbf{0.419} & 0.389 \\
    \bottomrule
    \end{tabular}
    }

    \label{tab:abl_4d_novel_views}
\end{table}

\section{\textbf{Editing 3D proxy with automatic propagation}}
\label{sec:supp_step2}

\subsection{Implementation Details}

Our approach supports diverse video editing tasks, including but not limited to pose editing, rotation, scaling, translation, texture modification, and object composition. While the key idea and implementation rely on our dual-propagation strategy, specific task implementation may involve task-dependent settings and variations.

\noindent \textbf{General Editing.} 
For four types of editing, pose editing, rotation, translation, and scaling, users only need to apply their desired modifications to the canonical mesh once, after which the edits are automatically propagated across all frames.

\noindent \textbf{Object Composition.} 
To composite a static object into a dynamic object, we first generate the 3D mesh of the new object. The generated mesh is then imported into Blender, where it is manually positioned to align with the intended location, forming a new edited mesh. Next, an anchor vertex is selected from the canonical mesh of the original dynamic object by finding the closest vertex to the new object. The motion of this anchor vertex is then transferred to all vertex of the new object, ensuring consistent movement within the scene. Additionally, users can achieve more diverse motion effects by applying complex binding strategies to the new object.

\noindent \textbf{Texture modification.} 
Texture editing in this context refers to directly modifying the vertex color of the mesh (\eg, editing the flower in \figref{fig:all_results}). Since the canonical mesh does not store color information, we replace it with the mesh from the first frame (\ie, $t=0$). Users can modify the texture color using Blender’s `vertex paint’ function or employ advanced UV mapping techniques for more precise adjustments.

\section{\textbf{Generative rendering from 3D proxy}}
\label{sec:supp_step3}

\begin{figure*}[th!]
\includegraphics[width=0.95\textwidth]{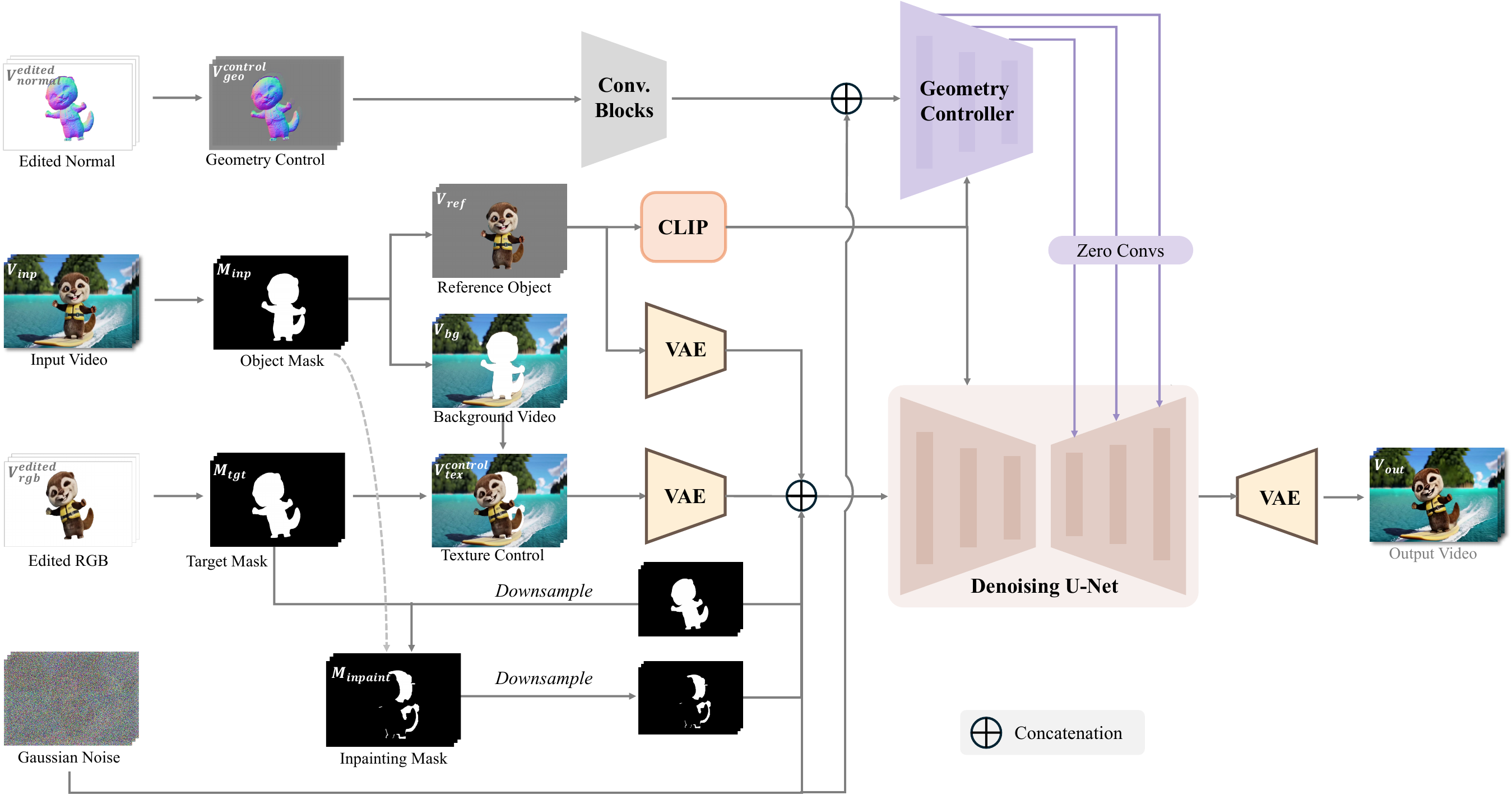}
  \caption{Inference pipeline of our decoupled video diffusion. }
  \label{fig:s3_workflow}
\end{figure*}

\subsection{Data Preparation}

As our decoupled video diffusion model operates in a self-supervised manner, we provide more details on the data construction process.

\noindent \textbf{Reference Object.} 
For a target object in a video, we first extract the object from the background using the ground-truth mask provided in the VOS dataset. Next, we apply random scaling, shifting, and rotation augmentations to all frames \textbf{simultaneously}, ensuring that the augmentation parameters remain consistent across all frames (including both the image and the corresponding mask). After augmentation, the background of the target object is filled with a uniform gray color. In addition, during the training process, we also randomly re-order the input video frames to simulate the reference object. 
In this way, we simulate the generation of pre-editing video data.

\noindent \textbf{Texture Control.} 
Since our dual-propagation strategy relies on nearest-neighbor mapping for texture propagation, the texture colors exhibit a locally similar pattern. To simulate this characteristic, we first apply Simple Linear Iterative Clustering (SLIC) to segment each frame into multiple small patches, with the number of patches randomly selected from the range [800, 1200]. Next, we use median blur filtering to smooth details within the patches and reduce boundary artifacts, with the kernel size randomly chosen from the range [3, 11]. Finally, consecutive downsampling and upsampling operations are applied, with the scaling factor randomly selected from the range [0.25, 0.5]. In this way, we simulate the coarse texture after editing.

\noindent \textbf{Geometry control.}We utilize Depth-Anything~\cite{yang2024depth}, to generate depth maps of the Ground-Truth object. These depth maps are then converted into normal maps, serving as the geometry control input for the geometry controller.

\subsection{Inference Workflow}

The inference pipeline is shown in \figref{fig:s3_workflow}. 
Formally, we denote the input video as $\mathbf{V}_{inp}$. The object mask for editing, obtained from SAM, is represented as $\mathbf{M}_{inp}$. 
The edited color is denoted by $\mathbf{V}^{edited}_{rgb}$, and the edited normal map is represented as $\mathbf{V}_{normal}^{edited}$. The target object mask is extracted by SAM and denoted by $\mathbf{M}_{tgt}$. 
We then obtain the inpainting mask $\mathbf{M}_{inpaint}$ by subtracting the overlapping regions of $\mathbf{M}_{inp}$ and $\mathbf{M}_{tgt}$ from $\mathbf{M}_{inp}$: 
\begin{equation}
    \mathbf{M}_{inpaint} = \mathbf{M}_{inp} - (\mathbf{M}_{inp} \cap \mathbf{M}_{tgt})
\end{equation}
To reduce the distraction from the backgrounds of the input video $\mathbf{V}_{inp}$, we mask its background and only retain its foreground object, to obtain the reference object video $\mathbf{V}_{ref}$: 
\begin{equation}
    \mathbf{V}_{ref} \text{,} \mathbf{V}_{bg} = \text{Split}(\mathbf{V}_{inp} \text{,} \mathbf{M}_{inp} )
\end{equation}
where $\text{Split}(\cdot)$ denotes the binary separating function and $\mathbf{V}_{bg}$ represents the background information, in which the foreground regions in $\mathbf{V}_{bg}$ are filled with white color, and the background regions in $\mathbf{V}_{ref}$ are filled with gray color, respectively. 
Then, the texture control map can be obtained by:
\begin{equation}
    \mathbf{V}_{tex}^{control} = \mathbf{V}^{edited}_{rgb} \times \mathbf{M}_{tgt}  + \mathbf{V}_{bg} \times (1 - \mathbf{M}_{tgt})
\end{equation}
We then obtain the geometry control map $\mathbf{V}_{geo}^{control} $ by replacing the background region in $\mathbf{V}^{edited}_{normal}$ with a gray color. 

With all inputs ready, we first convert the reference object $\mathbf{V}_{ref}$ and the texture control $\mathbf{V}_{tex}^{control}$ from pixel-space to latent space using VAE's encoder. 
We then apply nearest-neighbor down-sampling to the target mask $\mathbf{M}_{tgt}$ and the inpainting mask $\mathbf{M}_{inpaint}$ to resize them.
For geometry control, we first extract its features using several consecutive convolutional blocks with down-sampling~\cite{zhang2023adding}. These extracted features are then concatenated with the noisy latents obtained from the base model and passed into the control branch.
The CLIP image embeddings extracted from the reference object $\mathbf{V}_{ref}$ are also used as Key and Value in the cross-attention of the base model and the control branch. 
Finally, the denoised latents are sent to the VAE decoder to generate the output editing results.

\subsection{Implementation Details}
We use the I2V release of the stable video diffusion (SVD) \cite{blattmann2023stable} as the base model. 
Our model is trained on the VOS \cite{xu2018youtube} dataset~\footnote{Note that all examples used during testing are entirely unseen during training.}, filtering out extremely small-sized objects from the original 7800 unique objects to retain 5968 samples. 
As we do not have real paired data (source/target videos), we generate a training dataset via on-the-fly augmentation (Supp. C.1). 
We use a two-stage mixed-training method, first training the geometry controller and then fixing it to train the denoising UNet. 
In both stages, we adopt the same denoising loss as in SVD. 
The two training stages are optimized for 120K and 60K iterations, respectively, using the Adam \cite{loshchilov2017decoupled} optimizer on 8 NVIDIA A100 GPUs for 7 days. 
Each GPU processes a batch size of 1 with a resolution of 384×256. During testing, the default resolution is $768\times512$, but the model supports higher resolutions. 
During training, classifier-free guidance is applied to the reference object with a probability of 0.1. 
During the inference, we use the Euler \cite{karras2022elucidating} sampler with 25 sampling steps and a classifier-free guidance scale of 3.

\subsection{Experimental Details}
\noindent \textbf{Metrics.} We adopt the widely-used video editing evaluation metrics: Fram-Acc and Tem-Con. 
Fram-Acc measures the frame-wise editing accuracy, defined as the percentage of frames where the edited image achieves a higher CLIP similarity to the target prompt than to the source prompt. 
Tem-Con evaluates temporal consistency by calculating the cosine similarity between consecutive frame pairs. 
However, for fine-grained edit types, CLIP cannot measure the consistency of objects before and after editing. To this end, we introduce a new metric, CLAPScore (CLIP-APpearance Score), to jointly consider textual and semantic consistency. 
CLAPScore measures the accumulative error of the Fram-Acc and the DINO similarity score: 
\[
\text{CLAP Score} = \text{Fram-Acc} \times \text{DINO}(input, output),
\]
where $input$ and $output$ represent the input and edited frames.

\noindent \textbf{\dataset.} To evaluate the fine-grained video editing capabilities, we collect a new benchmark dataset. The videos in this dataset are sourced from the Internet \cite{pexels}, the DAVIS dataset \cite{perazzi2016benchmark}, and generated videos \cite{chai2023stablevideo, videoworldsimulators2024}. 
This dataset includes diverse video content across categories such as animals, humans, vehicles, and \etc, covering six types of video editing tasks: pose editing, rotation, scaling, translation, texture modification, and object composition. 
As many video editing methods rely on text prompts, we leverage GPT-4o to generate source, target, and instruction-based prompts by providing the input keyframe along with editing details. 
Specifically, we first manually edit a keyframe, typically the first frame, using ImageSculpturing~\cite{yenphraphai2024image}, which serves as the edited reference. The original and edited frames are then input to GPT-4o, accompanied by specific editing instructions to generate appropriate prompts.
The prompt to GPT-4o is: ``\textit{Describe the difference between the two images before and after the object is edited. Note that you should describe the object and its state. You do not need to mention phrases like “object xx does not exist” or “no operation was done” in the source prompt; simply describe the state of the edited object in the target prompt. Avoid using vague or redundant phrases, such as “in a simple and unmodified state.”  The editing for this case is [xxx]}''. 
We will release this benchmark dataset.

\noindent \textbf{User Study.}
In addition to measuring high-level input-output similarity, we conduct a user study to evaluate the perceptual quality of our approach in two aspects: editing quality (EQ), semantic consistency (SC), and visual plausibility (VP). We collect results from four methods compared and from our own on the \dataset benchmark. The videos were randomized and presented to 45 participants. 
For editing quality, participants ranked the results based on their alignment with the edited 3D mesh, with higher alignment receiving a higher rank. For semantic consistency, rankings were based on the degree of variation in the object before and after editing, where greater variation resulted in a lower rank. 
For visual plausibility, participants are asked to evaluate the editing realism by considering the overall visual realism, lighting consistency, and geometric coherence.
The results in Table 1 of the main paper indicate that our method consistently outperforms others and is the most preferred.

\section{\textbf{Discussion}}
\label{sec:supp_discussions}

\subsection{Pipeline Design}

While one might initially perceive our framework as an intricate integration of multiple components, it is designed explicitly to address several crucial challenges in 3D-aware video editing through a modular structure. 
Specifically, our method introduces: (1) a novel-view augmentation strategy combined with balanced-view sampling to significantly improve 3D proxy reconstruction consistency; 
(2) a dual-propagation strategy utilizing canonical Gaussian and mesh representations to efficiently propagate geometry and texture edits without the need for frame-by-frame manual adjustments; 
and (3) a decoupled video diffusion model trained with a self-supervised mixed-training strategy to ensure appearance consistency, conditioned directly on the geometry and texture from the manipulated 3D proxies. 
This clearly structured pipeline—consisting of distinct stages for reconstruction, interactive manipulation, and generative rendering—facilitates ease of use, maintainability, and future extensibility. 
Therefore, the perceived complexity is actually strategic, enhancing the robustness and flexibility of our approach (\eg, swapping in improved reconstruction or diffusion modules), and the integration of these innovative elements is what enables our novel and effective solution.

\begin{table}[t!]
\centering
\caption{Inference time comparison between our method and state-of-the-art (SOTA) methods.
To ensure a fair comparison with I2V-Edit, we first apply Image-Sculpting to edit the initial frame and then use the resulting edited frame as input for the subsequent process.}
\setlength{\tabcolsep}{4pt} 
\scalebox{0.9}{ 
\begin{tabular}{l|ccccc}
\toprule

 Methods &Tune-a-video & Pix2Video & I2V-Edit & Drag-Video  & Ours \\
\midrule
 Time (mins) & 10.7  & 2.8 & 44.3  & 12.2   & 91.4    \\

\bottomrule
\end{tabular}}
\label{tab:speed}
\end{table}

\begin{figure*}[h]
  \includegraphics[width=1\textwidth]{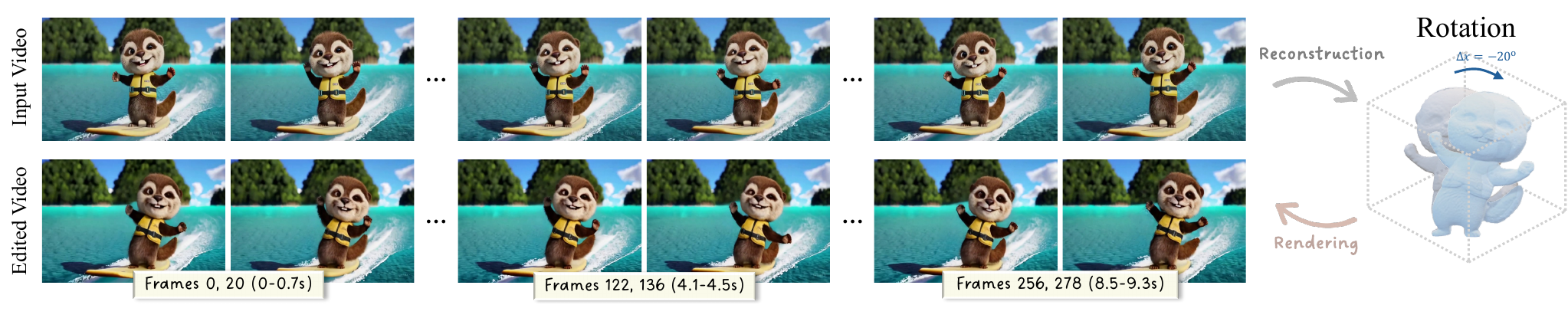}
  \vspace{-4mm}
  \caption{A 9.3-second edited video example demonstrating our method’s ability to handle long video sequences.}
  \label{fig:results_long}
\end{figure*}

\subsection{Efficiency Comparisons}
We conducted the inference speed comparisons at a fixed resolution (512×512) for a 14-frame video using a single A100 GPU. The result is shown in \tableref{tab:speed}. Note that we exclude manual editing time, which naturally varies based on the complexity and user interaction.
Although our framework requires relatively more computational time, with approximately 91 minutes for reconstruction and 43 seconds for the video diffusion model stage, this trade-off is justified by significantly improved geometric fidelity and temporally coherent appearances, as demonstrated in our experiments (Tab. 1 \& Fig. 6 of the main paper). 
Moreover, the design of our pipeline emphasizes practical usability: once the initial 3D proxy reconstruction is completed, multiple subsequent edits can be efficiently performed without re-optimization (\eg, rotate the object or adjust its pose), unlike most alternatives, which require a full recomputation for each new edit. 
Thus, despite its higher initial reconstruction time, our approach provides substantial long-term efficiency benefits and enhanced editing quality, aligning closely with practical usage scenarios where precision and consistency are paramount. 
We also believe that future developments in faster 4D reconstruction methods for video will further reduce the computational overhead, further improving the practicality of our approach.

\subsection{Topology of Extracted Mesh}

\noindent \textbf{Topology and Quality of Extracted Meshes.} A potential concern in the propagation of geometry is that extracted meshes might exhibit topological inconsistencies, such as disconnected components or internal holes. To address this, we enhance mesh quality by reducing geometric inconsistencies through balanced-view-sampling (BVS) and a scale-invariant depth constraint. These strategies help mitigate artifacts such as sunken surfaces (holes) and floating outliers. The effectiveness of these measures is validated through ablation studies presented in Fig.7 of the main paper.

\noindent \textbf{Reconciling Topological Differences.} Meshes extracted at different temporal instances may differ in topology. Given that the number of Gaussian points remains fixed throughout propagation, we utilize Gaussian points as intermediaries to reconcile these topological discrepancies. Specifically, we impose mask, RGB, and depth constraints during reconstruction to align mesh vertices closely with Gaussian points, thereby ensuring consistent geometry representation.

\subsection{Baseline Reconstruction Performance}

In Fig.7 (row 1) of the main paper, the baseline configuration (without novel views) utilizes \textbf{only the input frames} directly for dynamic 3D reconstruction, optimizing solely the observed view across time steps. Consequently, this configuration yields inferior reconstruction quality for unseen viewpoints due to its inherent lack of multi-view constraints. While state-of-the-art image-to-3D methods achieve impressive single-image reconstructions, their performance primarily focuses on static objects. These methods typically lack explicit inter-frame correspondence modeling, making them unsuitable for maintaining the temporal consistency necessary for coherent dynamic video editing tasks.

\subsection{Influence of the 3D Proxy Quality}

The reconstructed geometry and texture from the rendered 3D proxy serve as critical control signals guiding the video generation process. The quality of this proxy can impact the effectiveness and coherence of subsequent editing tasks. For instance, as illustrated in Fig.9 and quantified in Tab.3 of the main paper, removing texture controls (\ie, relying only on stage-1 geometry control) results in notably diminished visual quality. This degradation is primarily due to limited data augmentation during model training, as well as spatial misalignment between the target object before and after editing. Thus, inaccuracies or lower fidelity in the proxy’s geometry and texture inherently propagate through the pipeline, leading to compromised appearance and coherence in the final generated video. Incorporating more advanced or robust video reconstruction methods capable of providing higher-quality proxies could potentially enhance the overall coherence and controllability of the generated video, opening avenues for further improvements in dynamic editing tasks.

\subsection{User APIs}
Our framework uses Blender's intuitive and widely-used API along with automatic rigging tools (\eg, Mixamo\footnote{https://www.mixamo.com/} and Anything World\footnote{https://anything.world/}) to streamline the mesh manipulation. 
For instance, once the canonical mesh is reconstructed, we can first apply auto-rigging to generate a skeletal structure. Subsequently, the geometry of the object can be easily controlled by manipulating skeletal keypoints.

\subsection{Long Video Processing}

In the video diffusion stage, to handle sequences exceeding the default 14-frame limitation of the SVD model, we divide the video into several overlapping windows. The overlapping regions from previous windows are utilized to initialize the noise of subsequent windows via an SDEdit~\cite{meng2021sdedit}-based initialization strategy. Then, overlapping segments across windows are seamlessly merged using a progressive alpha-blending mechanism, where the weight of the previous window gradually decreases from 1 to 0, while the weight of the next window correspondingly increases from 0 to 1 across overlapping frames, ensuring smooth temporal transitions. 
In \figref{fig:results_long}, we present an example of editing results on a 9.3-second video. The result exhibits consistent geometry and appearance throughout, highlighting temporal coherence over extended durations.

\section{Additional Visual Results}
\label{sec:supp_visual}
\begin{figure*}[h!]
  \includegraphics[width=0.95\textwidth]{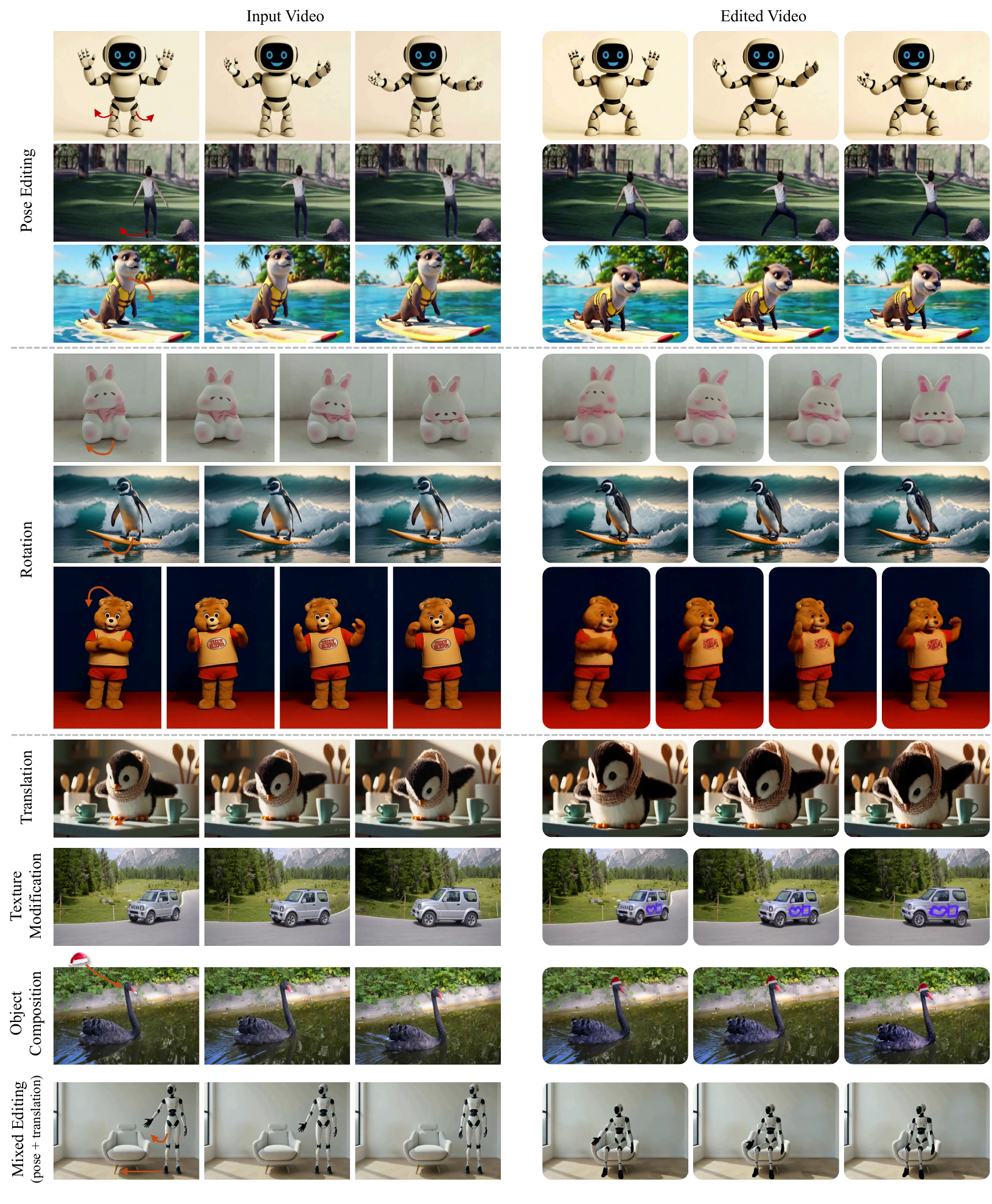}
  \vspace{-4mm}
  \caption{Additional visual results of our method. For each case, three or four input frames are shown on the left, and the edited results are displayed on the right. For each case, we highlight the editing using the red arrow in the first frame, except the texture modification.}
  \label{fig:all_results}
\end{figure*}

We present additional visual results of our approach in \figref{fig:all_results}, showcasing various editing types, including pose editing, rotation, translation, texture modification, object composition, and mixed edits (\ie, combinations of different editing types).

\newpage
\newpage

\bibliographystyle{ACM-Reference-Format}
\bibliography{sample-base}


\begin{thebibliography}{76}


\ifx \showCODEN    \undefined \def \showCODEN     #1{\unskip}     \fi
\ifx \showDOI      \undefined \def \showDOI       #1{#1}\fi
\ifx \showISBNx    \undefined \def \showISBNx     #1{\unskip}     \fi
\ifx \showISBNxiii \undefined \def \showISBNxiii  #1{\unskip}     \fi
\ifx \showISSN     \undefined \def \showISSN      #1{\unskip}     \fi
\ifx \showLCCN     \undefined \def \showLCCN      #1{\unskip}     \fi
\ifx \shownote     \undefined \def \shownote      #1{#1}          \fi
\ifx \showarticletitle \undefined \def \showarticletitle #1{#1}   \fi
\ifx \showURL      \undefined \def \showURL       {\relax}        \fi
\providecommand\bibfield[2]{#2}
\providecommand\bibinfo[2]{#2}
\providecommand\natexlab[1]{#1}
\providecommand\showeprint[2][]{arXiv:#2}

\bibitem[Achanta et~al\mbox{.}(2012)]%
        {achanta2012slic}
\bibfield{author}{\bibinfo{person}{Radhakrishna Achanta}, \bibinfo{person}{Appu Shaji}, \bibinfo{person}{Kevin Smith}, \bibinfo{person}{Aurelien Lucchi}, \bibinfo{person}{Pascal Fua}, {and} \bibinfo{person}{Sabine S{\"u}sstrunk}.} \bibinfo{year}{2012}\natexlab{}.
\newblock \showarticletitle{SLIC superpixels compared to state-of-the-art superpixel methods}.
\newblock \bibinfo{journal}{\emph{IEEE transactions on pattern analysis and machine intelligence}} \bibinfo{volume}{34}, \bibinfo{number}{11} (\bibinfo{year}{2012}), \bibinfo{pages}{2274--2282}.
\newblock


\bibitem[Bahmani et~al\mbox{.}(2024)]%
        {bahmani2024tc4d}
\bibfield{author}{\bibinfo{person}{Sherwin Bahmani}, \bibinfo{person}{Xian Liu}, \bibinfo{person}{Wang Yifan}, \bibinfo{person}{Ivan Skorokhodov}, \bibinfo{person}{Victor Rong}, \bibinfo{person}{Ziwei Liu}, \bibinfo{person}{Xihui Liu}, \bibinfo{person}{Jeong~Joon Park}, \bibinfo{person}{Sergey Tulyakov}, \bibinfo{person}{Gordon Wetzstein}, {et~al\mbox{.}}} \bibinfo{year}{2024}\natexlab{}.
\newblock \showarticletitle{Tc4d: Trajectory-conditioned text-to-4d generation}. In \bibinfo{booktitle}{\emph{European Conference on Computer Vision}}. Springer, \bibinfo{pages}{53--72}.
\newblock


\bibitem[Bar-Tal et~al\mbox{.}(2022)]%
        {bar2022text2live}
\bibfield{author}{\bibinfo{person}{Omer Bar-Tal}, \bibinfo{person}{Dolev Ofri-Amar}, \bibinfo{person}{Rafail Fridman}, \bibinfo{person}{Yoni Kasten}, {and} \bibinfo{person}{Tali Dekel}.} \bibinfo{year}{2022}\natexlab{}.
\newblock \showarticletitle{Text2live: Text-driven layered image and video editing}. In \bibinfo{booktitle}{\emph{European conference on computer vision}}. Springer, \bibinfo{pages}{707--723}.
\newblock


\bibitem[Bian et~al\mbox{.}(2025)]%
        {bian2025gs}
\bibfield{author}{\bibinfo{person}{Weikang Bian}, \bibinfo{person}{Zhaoyang Huang}, \bibinfo{person}{Xiaoyu Shi}, \bibinfo{person}{Yijin Li}, \bibinfo{person}{Fu-Yun Wang}, {and} \bibinfo{person}{Hongsheng Li}.} \bibinfo{year}{2025}\natexlab{}.
\newblock \showarticletitle{Gs-dit: Advancing video generation with pseudo 4d gaussian fields through efficient dense 3d point tracking}.
\newblock \bibinfo{journal}{\emph{arXiv preprint arXiv:2501.02690}} (\bibinfo{year}{2025}).
\newblock


\bibitem[Blattmann et~al\mbox{.}(2023)]%
        {blattmann2023stable}
\bibfield{author}{\bibinfo{person}{Andreas Blattmann}, \bibinfo{person}{Tim Dockhorn}, \bibinfo{person}{Sumith Kulal}, \bibinfo{person}{Daniel Mendelevitch}, \bibinfo{person}{Maciej Kilian}, \bibinfo{person}{Dominik Lorenz}, \bibinfo{person}{Yam Levi}, \bibinfo{person}{Zion English}, \bibinfo{person}{Vikram Voleti}, \bibinfo{person}{Adam Letts}, {et~al\mbox{.}}} \bibinfo{year}{2023}\natexlab{}.
\newblock \showarticletitle{Stable video diffusion: Scaling latent video diffusion models to large datasets}.
\newblock \bibinfo{journal}{\emph{arXiv preprint arXiv:2311.15127}} (\bibinfo{year}{2023}).
\newblock


\bibitem[Brooks et~al\mbox{.}(2024)]%
        {videoworldsimulators2024}
\bibfield{author}{\bibinfo{person}{Tim Brooks}, \bibinfo{person}{Bill Peebles}, \bibinfo{person}{Connor Holmes}, \bibinfo{person}{Will DePue}, \bibinfo{person}{Yufei Guo}, \bibinfo{person}{Li Jing}, \bibinfo{person}{David Schnurr}, \bibinfo{person}{Joe Taylor}, \bibinfo{person}{Troy Luhman}, \bibinfo{person}{Eric Luhman}, \bibinfo{person}{Clarence Ng}, \bibinfo{person}{Ricky Wang}, {and} \bibinfo{person}{Aditya Ramesh}.} \bibinfo{year}{2024}\natexlab{}.
\newblock \showarticletitle{Video generation models as world simulators}.
\newblock  (\bibinfo{year}{2024}).
\newblock
\urldef\tempurl%
\url{https://openai.com/research/video-generation-models-as-world-simulators}
\showURL{%
\tempurl}


\bibitem[Cai et~al\mbox{.}(2024)]%
        {cai2024generative}
\bibfield{author}{\bibinfo{person}{Shengqu Cai}, \bibinfo{person}{Duygu Ceylan}, \bibinfo{person}{Matheus Gadelha}, \bibinfo{person}{Chun-Hao~Paul Huang}, \bibinfo{person}{Tuanfeng~Yang Wang}, {and} \bibinfo{person}{Gordon Wetzstein}.} \bibinfo{year}{2024}\natexlab{}.
\newblock \showarticletitle{Generative rendering: Controllable 4d-guided video generation with 2d diffusion models}. In \bibinfo{booktitle}{\emph{Proceedings of the IEEE/CVF Conference on Computer Vision and Pattern Recognition}}. \bibinfo{pages}{7611--7620}.
\newblock


\bibitem[Ceylan et~al\mbox{.}(2023)]%
        {ceylan2023pix2video}
\bibfield{author}{\bibinfo{person}{Duygu Ceylan}, \bibinfo{person}{Chun-Hao~P Huang}, {and} \bibinfo{person}{Niloy~J Mitra}.} \bibinfo{year}{2023}\natexlab{}.
\newblock \showarticletitle{Pix2video: Video editing using image diffusion}. In \bibinfo{booktitle}{\emph{Proceedings of the IEEE/CVF International Conference on Computer Vision}}. \bibinfo{pages}{23206--23217}.
\newblock


\bibitem[Chai et~al\mbox{.}(2023)]%
        {chai2023stablevideo}
\bibfield{author}{\bibinfo{person}{Wenhao Chai}, \bibinfo{person}{Xun Guo}, \bibinfo{person}{Gaoang Wang}, {and} \bibinfo{person}{Yan Lu}.} \bibinfo{year}{2023}\natexlab{}.
\newblock \showarticletitle{Stablevideo: Text-driven consistency-aware diffusion video editing}. In \bibinfo{booktitle}{\emph{Proceedings of the IEEE/CVF International Conference on Computer Vision}}. \bibinfo{pages}{23040--23050}.
\newblock


\bibitem[Chen et~al\mbox{.}(2023)]%
        {chen2023control}
\bibfield{author}{\bibinfo{person}{Weifeng Chen}, \bibinfo{person}{Yatai Ji}, \bibinfo{person}{Jie Wu}, \bibinfo{person}{Hefeng Wu}, \bibinfo{person}{Pan Xie}, \bibinfo{person}{Jiashi Li}, \bibinfo{person}{Xin Xia}, \bibinfo{person}{Xuefeng Xiao}, {and} \bibinfo{person}{Liang Lin}.} \bibinfo{year}{2023}\natexlab{}.
\newblock \showarticletitle{Control-a-video: Controllable text-to-video generation with diffusion models}.
\newblock \bibinfo{journal}{\emph{arXiv preprint arXiv:2305.13840}} (\bibinfo{year}{2023}).
\newblock


\bibitem[Cheng et~al\mbox{.}(2025)]%
        {cheng20253d}
\bibfield{author}{\bibinfo{person}{Yen-Chi Cheng}, \bibinfo{person}{Krishna~Kumar Singh}, \bibinfo{person}{Jae~Shin Yoon}, \bibinfo{person}{Alexander Schwing}, \bibinfo{person}{Liangyan Gui}, \bibinfo{person}{Matheus Gadelha}, \bibinfo{person}{Paul Guerrero}, {and} \bibinfo{person}{Nanxuan Zhao}.} \bibinfo{year}{2025}\natexlab{}.
\newblock \showarticletitle{{3D-Fixup: Advancing Photo Editing with 3D Priors}}. In \bibinfo{booktitle}{\emph{Proceedings of the SIGGRAPH Conference Papers}}. \bibinfo{publisher}{ACM}.
\newblock
\showISBNx{979-8-4007-1540-2/2025/08}
\urldef\tempurl%
\url{https://doi.org/10.1145/3721238.3730695}
\showDOI{\tempurl}


\bibitem[Deng et~al\mbox{.}(2024)]%
        {deng2025dragvideo}
\bibfield{author}{\bibinfo{person}{Yufan Deng}, \bibinfo{person}{Ruida Wang}, \bibinfo{person}{Yuhao Zhang}, \bibinfo{person}{Yu-Wing Tai}, {and} \bibinfo{person}{Chi-Keung Tang}.} \bibinfo{year}{2024}\natexlab{}.
\newblock \showarticletitle{Dragvideo: Interactive drag-style video editing}. In \bibinfo{booktitle}{\emph{European Conference on Computer Vision}}. Springer, \bibinfo{pages}{183--199}.
\newblock


\bibitem[Esser et~al\mbox{.}(2023)]%
        {esser2023structure}
\bibfield{author}{\bibinfo{person}{Patrick Esser}, \bibinfo{person}{Johnathan Chiu}, \bibinfo{person}{Parmida Atighehchian}, \bibinfo{person}{Jonathan Granskog}, {and} \bibinfo{person}{Anastasis Germanidis}.} \bibinfo{year}{2023}\natexlab{}.
\newblock \showarticletitle{Structure and content-guided video synthesis with diffusion models}. In \bibinfo{booktitle}{\emph{Proceedings of the IEEE/CVF International Conference on Computer Vision}}.
\newblock


\bibitem[Fan et~al\mbox{.}(2024)]%
        {fan2024videoshop}
\bibfield{author}{\bibinfo{person}{Xiang Fan}, \bibinfo{person}{Anand Bhattad}, {and} \bibinfo{person}{Ranjay Krishna}.} \bibinfo{year}{2024}\natexlab{}.
\newblock \showarticletitle{Videoshop: Localized Semantic Video Editing with Noise-Extrapolated Diffusion Inversion}.
\newblock \bibinfo{journal}{\emph{arXiv preprint arXiv:2403.14617}} (\bibinfo{year}{2024}).
\newblock


\bibitem[Gu et~al\mbox{.}(2024)]%
        {gu2024videoswap}
\bibfield{author}{\bibinfo{person}{Yuchao Gu}, \bibinfo{person}{Yipin Zhou}, \bibinfo{person}{Bichen Wu}, \bibinfo{person}{Licheng Yu}, \bibinfo{person}{Jia-Wei Liu}, \bibinfo{person}{Rui Zhao}, \bibinfo{person}{Jay~Zhangjie Wu}, \bibinfo{person}{David~Junhao Zhang}, \bibinfo{person}{Mike~Zheng Shou}, {and} \bibinfo{person}{Kevin Tang}.} \bibinfo{year}{2024}\natexlab{}.
\newblock \showarticletitle{Videoswap: Customized video subject swapping with interactive semantic point correspondence}. In \bibinfo{booktitle}{\emph{Proceedings of the IEEE/CVF Conference on Computer Vision and Pattern Recognition}}. \bibinfo{pages}{7621--7630}.
\newblock


\bibitem[Gu et~al\mbox{.}(2025)]%
        {gu2025diffusion}
\bibfield{author}{\bibinfo{person}{Zekai Gu}, \bibinfo{person}{Rui Yan}, \bibinfo{person}{Jiahao Lu}, \bibinfo{person}{Peng Li}, \bibinfo{person}{Zhiyang Dou}, \bibinfo{person}{Chenyang Si}, \bibinfo{person}{Zhen Dong}, \bibinfo{person}{Qifeng Liu}, \bibinfo{person}{Cheng Lin}, \bibinfo{person}{Ziwei Liu}, {et~al\mbox{.}}} \bibinfo{year}{2025}\natexlab{}.
\newblock \showarticletitle{Diffusion as Shader: 3D-aware Video Diffusion for Versatile Video Generation Control}.
\newblock \bibinfo{journal}{\emph{arXiv preprint arXiv:2501.03847}} (\bibinfo{year}{2025}).
\newblock


\bibitem[Guo et~al\mbox{.}(2024b)]%
        {guo2024i2v}
\bibfield{author}{\bibinfo{person}{Xun Guo}, \bibinfo{person}{Mingwu Zheng}, \bibinfo{person}{Liang Hou}, \bibinfo{person}{Yuan Gao}, \bibinfo{person}{Yufan Deng}, \bibinfo{person}{Pengfei Wan}, \bibinfo{person}{Di Zhang}, \bibinfo{person}{Yufan Liu}, \bibinfo{person}{Weiming Hu}, \bibinfo{person}{Zhengjun Zha}, {et~al\mbox{.}}} \bibinfo{year}{2024}\natexlab{b}.
\newblock \showarticletitle{I2v-adapter: A general image-to-video adapter for diffusion models}. In \bibinfo{booktitle}{\emph{ACM SIGGRAPH 2024 Conference Papers}}.
\newblock


\bibitem[Guo et~al\mbox{.}(2024a)]%
        {guo2023animatediff}
\bibfield{author}{\bibinfo{person}{Yuwei Guo}, \bibinfo{person}{Ceyuan Yang}, \bibinfo{person}{Anyi Rao}, \bibinfo{person}{Zhengyang Liang}, \bibinfo{person}{Yaohui Wang}, \bibinfo{person}{Yu Qiao}, \bibinfo{person}{Maneesh Agrawala}, \bibinfo{person}{Dahua Lin}, {and} \bibinfo{person}{Bo Dai}.} \bibinfo{year}{2024}\natexlab{a}.
\newblock \showarticletitle{AnimateDiff: Animate Your Personalized Text-to-Image Diffusion Models without Specific Tuning}.
\newblock \bibinfo{journal}{\emph{International Conference on Learning Representations}} (\bibinfo{year}{2024}).
\newblock


\bibitem[Ho et~al\mbox{.}(2022)]%
        {ho2022imagen}
\bibfield{author}{\bibinfo{person}{Jonathan Ho}, \bibinfo{person}{William Chan}, \bibinfo{person}{Chitwan Saharia}, \bibinfo{person}{Jay Whang}, \bibinfo{person}{Ruiqi Gao}, \bibinfo{person}{Alexey Gritsenko}, \bibinfo{person}{Diederik~P Kingma}, \bibinfo{person}{Ben Poole}, \bibinfo{person}{Mohammad Norouzi}, \bibinfo{person}{David~J Fleet}, {et~al\mbox{.}}} \bibinfo{year}{2022}\natexlab{}.
\newblock \showarticletitle{Imagen video: High definition video generation with diffusion models}.
\newblock \bibinfo{journal}{\emph{arXiv preprint arXiv:2210.02303}} (\bibinfo{year}{2022}).
\newblock


\bibitem[Ho et~al\mbox{.}(2020)]%
        {ho2020denoising}
\bibfield{author}{\bibinfo{person}{Jonathan Ho}, \bibinfo{person}{Ajay Jain}, {and} \bibinfo{person}{Pieter Abbeel}.} \bibinfo{year}{2020}\natexlab{}.
\newblock \showarticletitle{Denoising diffusion probabilistic models}.
\newblock \bibinfo{journal}{\emph{Advances in neural information processing systems}}  \bibinfo{volume}{33} (\bibinfo{year}{2020}), \bibinfo{pages}{6840--6851}.
\newblock


\bibitem[Hu and Xu(2023)]%
        {hu2023videocontrolnet}
\bibfield{author}{\bibinfo{person}{Zhihao Hu} {and} \bibinfo{person}{Dong Xu}.} \bibinfo{year}{2023}\natexlab{}.
\newblock \showarticletitle{Videocontrolnet: A motion-guided video-to-video translation framework by using diffusion model with controlnet}.
\newblock \bibinfo{journal}{\emph{arXiv preprint arXiv:2307.14073}} (\bibinfo{year}{2023}).
\newblock


\bibitem[Huang et~al\mbox{.}(2025)]%
        {huang2025voyager}
\bibfield{author}{\bibinfo{person}{Tianyu Huang}, \bibinfo{person}{Wangguandong Zheng}, \bibinfo{person}{Tengfei Wang}, \bibinfo{person}{Yuhao Liu}, \bibinfo{person}{Zhenwei Wang}, \bibinfo{person}{Junta Wu}, \bibinfo{person}{Jie Jiang}, \bibinfo{person}{Hui Li}, \bibinfo{person}{Rynson~WH Lau}, \bibinfo{person}{Wangmeng Zuo}, {et~al\mbox{.}}} \bibinfo{year}{2025}\natexlab{}.
\newblock \showarticletitle{Voyager: Long-Range and World-Consistent Video Diffusion for Explorable 3D Scene Generation}.
\newblock \bibinfo{journal}{\emph{arXiv preprint arXiv:2506.04225}} (\bibinfo{year}{2025}).
\newblock


\bibitem[Jiang et~al\mbox{.}(2024)]%
        {jiang2024animate3d}
\bibfield{author}{\bibinfo{person}{Yanqin Jiang}, \bibinfo{person}{Chaohui Yu}, \bibinfo{person}{Chenjie Cao}, \bibinfo{person}{Fan Wang}, \bibinfo{person}{Weiming Hu}, {and} \bibinfo{person}{Jin Gao}.} \bibinfo{year}{2024}\natexlab{}.
\newblock \showarticletitle{Animate3d: Animating any 3d model with multi-view video diffusion}.
\newblock \bibinfo{journal}{\emph{arXiv preprint arXiv:2407.11398}} (\bibinfo{year}{2024}).
\newblock


\bibitem[Kagaya et~al\mbox{.}(2010)]%
        {kagaya2010video}
\bibfield{author}{\bibinfo{person}{Mizuki Kagaya}, \bibinfo{person}{William Brendel}, \bibinfo{person}{Qingqing Deng}, \bibinfo{person}{Todd Kesterson}, \bibinfo{person}{Sinisa Todorovic}, \bibinfo{person}{Patrick~J Neill}, {and} \bibinfo{person}{Eugene Zhang}.} \bibinfo{year}{2010}\natexlab{}.
\newblock \showarticletitle{Video painting with space-time-varying style parameters}.
\newblock \bibinfo{journal}{\emph{IEEE transactions on visualization and computer graphics}} \bibinfo{volume}{17}, \bibinfo{number}{1} (\bibinfo{year}{2010}), \bibinfo{pages}{74--87}.
\newblock


\bibitem[Karras et~al\mbox{.}(2022)]%
        {karras2022elucidating}
\bibfield{author}{\bibinfo{person}{Tero Karras}, \bibinfo{person}{Miika Aittala}, \bibinfo{person}{Timo Aila}, {and} \bibinfo{person}{Samuli Laine}.} \bibinfo{year}{2022}\natexlab{}.
\newblock \showarticletitle{Elucidating the design space of diffusion-based generative models}.
\newblock \bibinfo{journal}{\emph{Advances in neural information processing systems}}  \bibinfo{volume}{35} (\bibinfo{year}{2022}), \bibinfo{pages}{26565--26577}.
\newblock


\bibitem[Kasten et~al\mbox{.}(2021)]%
        {kasten2021layered}
\bibfield{author}{\bibinfo{person}{Yoni Kasten}, \bibinfo{person}{Dolev Ofri}, \bibinfo{person}{Oliver Wang}, {and} \bibinfo{person}{Tali Dekel}.} \bibinfo{year}{2021}\natexlab{}.
\newblock \showarticletitle{Layered neural atlases for consistent video editing}.
\newblock \bibinfo{journal}{\emph{ACM Transactions on Graphics (TOG)}} \bibinfo{volume}{40}, \bibinfo{number}{6} (\bibinfo{year}{2021}), \bibinfo{pages}{1--12}.
\newblock


\bibitem[Kerbl et~al\mbox{.}(2023)]%
        {kerbl20233d}
\bibfield{author}{\bibinfo{person}{Bernhard Kerbl}, \bibinfo{person}{Georgios Kopanas}, \bibinfo{person}{Thomas Leimk{\"u}hler}, {and} \bibinfo{person}{George Drettakis}.} \bibinfo{year}{2023}\natexlab{}.
\newblock \showarticletitle{3d gaussian splatting for real-time radiance field rendering.}
\newblock \bibinfo{journal}{\emph{ACM Trans. Graph.}} \bibinfo{volume}{42}, \bibinfo{number}{4} (\bibinfo{year}{2023}), \bibinfo{pages}{139--1}.
\newblock


\bibitem[Koo et~al\mbox{.}(2025)]%
        {koo2025videohandles}
\bibfield{author}{\bibinfo{person}{Juil Koo}, \bibinfo{person}{Paul Guerrero}, \bibinfo{person}{Chun-Hao~P Huang}, \bibinfo{person}{Duygu Ceylan}, {and} \bibinfo{person}{Minhyuk Sung}.} \bibinfo{year}{2025}\natexlab{}.
\newblock \showarticletitle{Videohandles: Editing 3d object compositions in videos using video generative priors}. In \bibinfo{booktitle}{\emph{Proceedings of the Computer Vision and Pattern Recognition Conference}}. \bibinfo{pages}{17692--17701}.
\newblock


\bibitem[Ku et~al\mbox{.}(2024)]%
        {ku2024anyv2v}
\bibfield{author}{\bibinfo{person}{Max Ku}, \bibinfo{person}{Cong Wei}, \bibinfo{person}{Weiming Ren}, \bibinfo{person}{Huan Yang}, {and} \bibinfo{person}{Wenhu Chen}.} \bibinfo{year}{2024}\natexlab{}.
\newblock \showarticletitle{AnyV2V: A Tuning-Free Framework For Any Video-to-Video Editing Tasks}.
\newblock \bibinfo{journal}{\emph{Transactions on Machine Learning Research}} (\bibinfo{year}{2024}).
\newblock


\bibitem[Labs(2024)]%
        {flux2024}
\bibfield{author}{\bibinfo{person}{Black~Forest Labs}.} \bibinfo{year}{2024}\natexlab{}.
\newblock \bibinfo{title}{FLUX}.
\newblock \bibinfo{howpublished}{\url{https://github.com/black-forest-labs/flux}}.
\newblock


\bibitem[Laine et~al\mbox{.}(2020)]%
        {laine2020modular}
\bibfield{author}{\bibinfo{person}{Samuli Laine}, \bibinfo{person}{Janne Hellsten}, \bibinfo{person}{Tero Karras}, \bibinfo{person}{Yeongho Seol}, \bibinfo{person}{Jaakko Lehtinen}, {and} \bibinfo{person}{Timo Aila}.} \bibinfo{year}{2020}\natexlab{}.
\newblock \showarticletitle{Modular primitives for high-performance differentiable rendering}.
\newblock \bibinfo{journal}{\emph{ACM Transactions on Graphics (ToG)}} \bibinfo{volume}{39}, \bibinfo{number}{6} (\bibinfo{year}{2020}), \bibinfo{pages}{1--14}.
\newblock


\bibitem[Liang et~al\mbox{.}(2025)]%
        {liang2025vodiff}
\bibfield{author}{\bibinfo{person}{Dong Liang}, \bibinfo{person}{Jinyuan Jia}, \bibinfo{person}{Yuhao Liu}, \bibinfo{person}{Zhanghan Ke}, \bibinfo{person}{Hongbo Fu}, {and} \bibinfo{person}{Rynson~WH Lau}.} \bibinfo{year}{2025}\natexlab{}.
\newblock \showarticletitle{VODiff: Controlling Object Visibility Order in Text-to-Image Generation}. In \bibinfo{booktitle}{\emph{Proceedings of the Computer Vision and Pattern Recognition Conference}}. \bibinfo{pages}{18379--18389}.
\newblock


\bibitem[Liu et~al\mbox{.}(2024b)]%
        {liu2024dynamic}
\bibfield{author}{\bibinfo{person}{Isabella Liu}, \bibinfo{person}{Hao Su}, {and} \bibinfo{person}{Xiaolong Wang}.} \bibinfo{year}{2024}\natexlab{b}.
\newblock \showarticletitle{Dynamic Gaussians Mesh: Consistent Mesh Reconstruction from Monocular Videos}.
\newblock \bibinfo{journal}{\emph{arXiv preprint arXiv:2404.12379}} (\bibinfo{year}{2024}).
\newblock


\bibitem[Liu et~al\mbox{.}(2025)]%
        {liu2025generative}
\bibfield{author}{\bibinfo{person}{Shaoteng Liu}, \bibinfo{person}{Tianyu Wang}, \bibinfo{person}{Jui-Hsien Wang}, \bibinfo{person}{Qing Liu}, \bibinfo{person}{Zhifei Zhang}, \bibinfo{person}{Joon-Young Lee}, \bibinfo{person}{Yijun Li}, \bibinfo{person}{Bei Yu}, \bibinfo{person}{Zhe Lin}, \bibinfo{person}{Soo~Ye Kim}, {et~al\mbox{.}}} \bibinfo{year}{2025}\natexlab{}.
\newblock \showarticletitle{Generative video propagation}. In \bibinfo{booktitle}{\emph{Proceedings of the Computer Vision and Pattern Recognition Conference}}. \bibinfo{pages}{17712--17722}.
\newblock


\bibitem[Liu et~al\mbox{.}(2024a)]%
        {liu2024diff}
\bibfield{author}{\bibinfo{person}{Yuhao Liu}, \bibinfo{person}{Zhanghan Ke}, \bibinfo{person}{Fang Liu}, \bibinfo{person}{Nanxuan Zhao}, {and} \bibinfo{person}{Rynson~WH Lau}.} \bibinfo{year}{2024}\natexlab{a}.
\newblock \showarticletitle{Diff-plugin: Revitalizing details for diffusion-based low-level tasks}. In \bibinfo{booktitle}{\emph{Proceedings of the IEEE/CVF Conference on Computer Vision and Pattern Recognition}}. \bibinfo{pages}{4197--4208}.
\newblock


\bibitem[Loshchilov(2017)]%
        {loshchilov2017decoupled}
\bibfield{author}{\bibinfo{person}{I Loshchilov}.} \bibinfo{year}{2017}\natexlab{}.
\newblock \showarticletitle{Decoupled weight decay regularization}.
\newblock \bibinfo{journal}{\emph{arXiv preprint arXiv:1711.05101}} (\bibinfo{year}{2017}).
\newblock


\bibitem[Luiten et~al\mbox{.}(2023)]%
        {luiten2023dynamic}
\bibfield{author}{\bibinfo{person}{Jonathon Luiten}, \bibinfo{person}{Georgios Kopanas}, \bibinfo{person}{Bastian Leibe}, {and} \bibinfo{person}{Deva Ramanan}.} \bibinfo{year}{2023}\natexlab{}.
\newblock \showarticletitle{Dynamic 3d gaussians: Tracking by persistent dynamic view synthesis}.
\newblock \bibinfo{journal}{\emph{arXiv preprint arXiv:2308.09713}} (\bibinfo{year}{2023}).
\newblock


\bibitem[Lv et~al\mbox{.}(2024)]%
        {lv2024gpt4motion}
\bibfield{author}{\bibinfo{person}{Jiaxi Lv}, \bibinfo{person}{Yi Huang}, \bibinfo{person}{Mingfu Yan}, \bibinfo{person}{Jiancheng Huang}, \bibinfo{person}{Jianzhuang Liu}, \bibinfo{person}{Yifan Liu}, \bibinfo{person}{Yafei Wen}, \bibinfo{person}{Xiaoxin Chen}, {and} \bibinfo{person}{Shifeng Chen}.} \bibinfo{year}{2024}\natexlab{}.
\newblock \showarticletitle{GPT4Motion: Scripting Physical Motions in Text-to-Video Generation via Blender-Oriented GPT Planning}. In \bibinfo{booktitle}{\emph{Proceedings of the IEEE/CVF Conference on Computer Vision and Pattern Recognition}}. \bibinfo{pages}{1430--1440}.
\newblock


\bibitem[Ma et~al\mbox{.}(2023)]%
        {ma2023magicstick}
\bibfield{author}{\bibinfo{person}{Yue Ma}, \bibinfo{person}{Xiaodong Cun}, \bibinfo{person}{Yingqing He}, \bibinfo{person}{Chenyang Qi}, \bibinfo{person}{Xintao Wang}, \bibinfo{person}{Ying Shan}, \bibinfo{person}{Xiu Li}, {and} \bibinfo{person}{Qifeng Chen}.} \bibinfo{year}{2023}\natexlab{}.
\newblock \showarticletitle{Magicstick: Controllable video editing via control handle transformations}.
\newblock \bibinfo{journal}{\emph{arXiv preprint arXiv:2312.03047}} (\bibinfo{year}{2023}).
\newblock


\bibitem[Meng et~al\mbox{.}(2021)]%
        {meng2021sdedit}
\bibfield{author}{\bibinfo{person}{Chenlin Meng}, \bibinfo{person}{Yutong He}, \bibinfo{person}{Yang Song}, \bibinfo{person}{Jiaming Song}, \bibinfo{person}{Jiajun Wu}, \bibinfo{person}{Jun-Yan Zhu}, {and} \bibinfo{person}{Stefano Ermon}.} \bibinfo{year}{2021}\natexlab{}.
\newblock \showarticletitle{Sdedit: Guided image synthesis and editing with stochastic differential equations}.
\newblock \bibinfo{journal}{\emph{arXiv preprint arXiv:2108.01073}} (\bibinfo{year}{2021}).
\newblock


\bibitem[Michel et~al\mbox{.}(2023)]%
        {michel2023object}
\bibfield{author}{\bibinfo{person}{Oscar Michel}, \bibinfo{person}{Anand Bhattad}, \bibinfo{person}{Eli VanderBilt}, \bibinfo{person}{Ranjay Krishna}, \bibinfo{person}{Aniruddha Kembhavi}, {and} \bibinfo{person}{Tanmay Gupta}.} \bibinfo{year}{2023}\natexlab{}.
\newblock \showarticletitle{Object 3dit: Language-guided 3d-aware image editing}.
\newblock \bibinfo{journal}{\emph{Advances in Neural Information Processing Systems}}  \bibinfo{volume}{36} (\bibinfo{year}{2023}), \bibinfo{pages}{3497--3516}.
\newblock


\bibitem[Mildenhall et~al\mbox{.}(2021)]%
        {mildenhall2021nerf}
\bibfield{author}{\bibinfo{person}{Ben Mildenhall}, \bibinfo{person}{Pratul~P Srinivasan}, \bibinfo{person}{Matthew Tancik}, \bibinfo{person}{Jonathan~T Barron}, \bibinfo{person}{Ravi Ramamoorthi}, {and} \bibinfo{person}{Ren Ng}.} \bibinfo{year}{2021}\natexlab{}.
\newblock \showarticletitle{Nerf: Representing scenes as neural radiance fields for view synthesis}.
\newblock \bibinfo{journal}{\emph{Commun. ACM}} \bibinfo{volume}{65}, \bibinfo{number}{1} (\bibinfo{year}{2021}), \bibinfo{pages}{99--106}.
\newblock


\bibitem[Mou et~al\mbox{.}(2024)]%
        {mou2024revideo}
\bibfield{author}{\bibinfo{person}{Chong Mou}, \bibinfo{person}{Mingdeng Cao}, \bibinfo{person}{Xintao Wang}, \bibinfo{person}{Zhaoyang Zhang}, \bibinfo{person}{Ying Shan}, {and} \bibinfo{person}{Jian Zhang}.} \bibinfo{year}{2024}\natexlab{}.
\newblock \showarticletitle{ReVideo: Remake a Video with Motion and Content Control}.
\newblock \bibinfo{journal}{\emph{arXiv preprint arXiv:2405.13865}} (\bibinfo{year}{2024}).
\newblock


\bibitem[Ouyang et~al\mbox{.}(2024)]%
        {ouyang2024i2vedit}
\bibfield{author}{\bibinfo{person}{Wenqi Ouyang}, \bibinfo{person}{Yi Dong}, \bibinfo{person}{Lei Yang}, \bibinfo{person}{Jianlou Si}, {and} \bibinfo{person}{Xingang Pan}.} \bibinfo{year}{2024}\natexlab{}.
\newblock \showarticletitle{I2VEdit: First-Frame-Guided Video Editing via Image-to-Video Diffusion Models}. In \bibinfo{booktitle}{\emph{SIGGRAPH Asia 2024 Conference Papers}}.
\newblock


\bibitem[Pan et~al\mbox{.}(2023)]%
        {pan2023drag}
\bibfield{author}{\bibinfo{person}{Xingang Pan}, \bibinfo{person}{Ayush Tewari}, \bibinfo{person}{Thomas Leimk{\"u}hler}, \bibinfo{person}{Lingjie Liu}, \bibinfo{person}{Abhimitra Meka}, {and} \bibinfo{person}{Christian Theobalt}.} \bibinfo{year}{2023}\natexlab{}.
\newblock \showarticletitle{Drag your gan: Interactive point-based manipulation on the generative image manifold}. In \bibinfo{booktitle}{\emph{ACM SIGGRAPH 2023 Conference Proceedings}}.
\newblock


\bibitem[Pandey et~al\mbox{.}(2024)]%
        {pandey2024diffusion}
\bibfield{author}{\bibinfo{person}{Karran Pandey}, \bibinfo{person}{Paul Guerrero}, \bibinfo{person}{Matheus Gadelha}, \bibinfo{person}{Yannick Hold-Geoffroy}, \bibinfo{person}{Karan Singh}, {and} \bibinfo{person}{Niloy~J Mitra}.} \bibinfo{year}{2024}\natexlab{}.
\newblock \showarticletitle{Diffusion handles enabling 3d edits for diffusion models by lifting activations to 3d}. In \bibinfo{booktitle}{\emph{Proceedings of the IEEE/CVF Conference on Computer Vision and Pattern Recognition}}. \bibinfo{pages}{7695--7704}.
\newblock


\bibitem[Peng et~al\mbox{.}(2021)]%
        {peng2021shape}
\bibfield{author}{\bibinfo{person}{Songyou Peng}, \bibinfo{person}{Chiyu Jiang}, \bibinfo{person}{Yiyi Liao}, \bibinfo{person}{Michael Niemeyer}, \bibinfo{person}{Marc Pollefeys}, {and} \bibinfo{person}{Andreas Geiger}.} \bibinfo{year}{2021}\natexlab{}.
\newblock \showarticletitle{Shape as points: A differentiable poisson solver}.
\newblock \bibinfo{journal}{\emph{Advances in Neural Information Processing Systems}}  \bibinfo{volume}{34} (\bibinfo{year}{2021}), \bibinfo{pages}{13032--13044}.
\newblock


\bibitem[Perazzi et~al\mbox{.}(2016)]%
        {perazzi2016benchmark}
\bibfield{author}{\bibinfo{person}{Federico Perazzi}, \bibinfo{person}{Jordi Pont-Tuset}, \bibinfo{person}{Brian McWilliams}, \bibinfo{person}{Luc Van~Gool}, \bibinfo{person}{Markus Gross}, {and} \bibinfo{person}{Alexander Sorkine-Hornung}.} \bibinfo{year}{2016}\natexlab{}.
\newblock \showarticletitle{A benchmark dataset and evaluation methodology for video object segmentation}. In \bibinfo{booktitle}{\emph{Proceedings of the IEEE conference on computer vision and pattern recognition}}. \bibinfo{pages}{724--732}.
\newblock


\bibitem[Pexels({[n.\,d.]})]%
        {pexels}
\bibfield{author}{\bibinfo{person}{Pexels}.} \bibinfo{year}{[n.\,d.]}\natexlab{}.
\newblock \bibinfo{title}{PEXELS}.
\newblock \bibinfo{howpublished}{\url{https://www.pexels.com}}.
\newblock


\bibitem[Pumarola et~al\mbox{.}(2021)]%
        {pumarola2021d}
\bibfield{author}{\bibinfo{person}{Albert Pumarola}, \bibinfo{person}{Enric Corona}, \bibinfo{person}{Gerard Pons-Moll}, {and} \bibinfo{person}{Francesc Moreno-Noguer}.} \bibinfo{year}{2021}\natexlab{}.
\newblock \showarticletitle{D-nerf: Neural radiance fields for dynamic scenes}. In \bibinfo{booktitle}{\emph{Proceedings of the IEEE/CVF Conference on Computer Vision and Pattern Recognition}}. \bibinfo{pages}{10318--10327}.
\newblock


\bibitem[Radford et~al\mbox{.}(2021)]%
        {radford2021learning}
\bibfield{author}{\bibinfo{person}{Alec Radford}, \bibinfo{person}{Jong~Wook Kim}, \bibinfo{person}{Chris Hallacy}, \bibinfo{person}{Aditya Ramesh}, \bibinfo{person}{Gabriel Goh}, \bibinfo{person}{Sandhini Agarwal}, \bibinfo{person}{Girish Sastry}, \bibinfo{person}{Amanda Askell}, \bibinfo{person}{Pamela Mishkin}, \bibinfo{person}{Jack Clark}, {et~al\mbox{.}}} \bibinfo{year}{2021}\natexlab{}.
\newblock \showarticletitle{Learning transferable visual models from natural language supervision}. In \bibinfo{booktitle}{\emph{ICML}}.
\newblock


\bibitem[Ravi et~al\mbox{.}(2024)]%
        {ravi2024sam}
\bibfield{author}{\bibinfo{person}{Nikhila Ravi}, \bibinfo{person}{Valentin Gabeur}, \bibinfo{person}{Yuan-Ting Hu}, \bibinfo{person}{Ronghang Hu}, \bibinfo{person}{Chaitanya Ryali}, \bibinfo{person}{Tengyu Ma}, \bibinfo{person}{Haitham Khedr}, \bibinfo{person}{Roman R{\"a}dle}, \bibinfo{person}{Chloe Rolland}, \bibinfo{person}{Laura Gustafson}, {et~al\mbox{.}}} \bibinfo{year}{2024}\natexlab{}.
\newblock \showarticletitle{Sam 2: Segment anything in images and videos}.
\newblock \bibinfo{journal}{\emph{arXiv preprint arXiv:2408.00714}} (\bibinfo{year}{2024}).
\newblock


\bibitem[Ren et~al\mbox{.}(2024)]%
        {ren2024l4gm}
\bibfield{author}{\bibinfo{person}{Jiawei Ren}, \bibinfo{person}{Cheng Xie}, \bibinfo{person}{Ashkan Mirzaei}, \bibinfo{person}{Karsten Kreis}, \bibinfo{person}{Ziwei Liu}, \bibinfo{person}{Antonio Torralba}, \bibinfo{person}{Sanja Fidler}, \bibinfo{person}{Seung~Wook Kim}, \bibinfo{person}{Huan Ling}, {et~al\mbox{.}}} \bibinfo{year}{2024}\natexlab{}.
\newblock \showarticletitle{L4gm: Large 4d gaussian reconstruction model}.
\newblock \bibinfo{journal}{\emph{Advances in Neural Information Processing Systems}}  \bibinfo{volume}{37} (\bibinfo{year}{2024}), \bibinfo{pages}{56828--56858}.
\newblock


\bibitem[Rombach et~al\mbox{.}(2022)]%
        {rombach2022high}
\bibfield{author}{\bibinfo{person}{Robin Rombach}, \bibinfo{person}{Andreas Blattmann}, \bibinfo{person}{Dominik Lorenz}, \bibinfo{person}{Patrick Esser}, {and} \bibinfo{person}{Bj{\"o}rn Ommer}.} \bibinfo{year}{2022}\natexlab{}.
\newblock \showarticletitle{High-resolution image synthesis with latent diffusion models}. In \bibinfo{booktitle}{\emph{CVPR}}.
\newblock


\bibitem[Shi et~al\mbox{.}(2023a)]%
        {shi2023zero123++}
\bibfield{author}{\bibinfo{person}{Ruoxi Shi}, \bibinfo{person}{Hansheng Chen}, \bibinfo{person}{Zhuoyang Zhang}, \bibinfo{person}{Minghua Liu}, \bibinfo{person}{Chao Xu}, \bibinfo{person}{Xinyue Wei}, \bibinfo{person}{Linghao Chen}, \bibinfo{person}{Chong Zeng}, {and} \bibinfo{person}{Hao Su}.} \bibinfo{year}{2023}\natexlab{a}.
\newblock \showarticletitle{Zero123++: a single image to consistent multi-view diffusion base model}.
\newblock \bibinfo{journal}{\emph{arXiv preprint arXiv:2310.15110}} (\bibinfo{year}{2023}).
\newblock


\bibitem[Shi et~al\mbox{.}(2024)]%
        {shi2024motion}
\bibfield{author}{\bibinfo{person}{Xiaoyu Shi}, \bibinfo{person}{Zhaoyang Huang}, \bibinfo{person}{Fu-Yun Wang}, \bibinfo{person}{Weikang Bian}, \bibinfo{person}{Dasong Li}, \bibinfo{person}{Yi Zhang}, \bibinfo{person}{Manyuan Zhang}, \bibinfo{person}{Ka~Chun Cheung}, \bibinfo{person}{Simon See}, \bibinfo{person}{Hongwei Qin}, {et~al\mbox{.}}} \bibinfo{year}{2024}\natexlab{}.
\newblock \showarticletitle{Motion-i2v: Consistent and controllable image-to-video generation with explicit motion modeling}. In \bibinfo{booktitle}{\emph{ACM SIGGRAPH 2024 Conference Papers}}. \bibinfo{pages}{1--11}.
\newblock


\bibitem[Shi et~al\mbox{.}(2023b)]%
        {shi2023mvdream}
\bibfield{author}{\bibinfo{person}{Yichun Shi}, \bibinfo{person}{Peng Wang}, \bibinfo{person}{Jianglong Ye}, \bibinfo{person}{Mai Long}, \bibinfo{person}{Kejie Li}, {and} \bibinfo{person}{Xiao Yang}.} \bibinfo{year}{2023}\natexlab{b}.
\newblock \showarticletitle{Mvdream: Multi-view diffusion for 3d generation}.
\newblock \bibinfo{journal}{\emph{arXiv preprint arXiv:2308.16512}} (\bibinfo{year}{2023}).
\newblock


\bibitem[Shin et~al\mbox{.}(2024)]%
        {shin2024enhancing}
\bibfield{author}{\bibinfo{person}{Inkyu Shin}, \bibinfo{person}{Qihang Yu}, \bibinfo{person}{Xiaohui Shen}, \bibinfo{person}{In~So Kweon}, \bibinfo{person}{Kuk-Jin Yoon}, {and} \bibinfo{person}{Liang-Chieh Chen}.} \bibinfo{year}{2024}\natexlab{}.
\newblock \showarticletitle{Enhancing Temporal Consistency in Video Editing by Reconstructing Videos with 3D Gaussian Splatting}.
\newblock \bibinfo{journal}{\emph{arXiv preprint arXiv:2406.02541}} (\bibinfo{year}{2024}).
\newblock


\bibitem[Song et~al\mbox{.}(2021)]%
        {song2021denoising}
\bibfield{author}{\bibinfo{person}{Jiaming Song}, \bibinfo{person}{Chenlin Meng}, {and} \bibinfo{person}{Stefano Ermon}.} \bibinfo{year}{2021}\natexlab{}.
\newblock \showarticletitle{Denoising Diffusion Implicit Models}. In \bibinfo{booktitle}{\emph{ICLR}}.
\newblock


\bibitem[Sun et~al\mbox{.}(2024)]%
        {sun2024splatter}
\bibfield{author}{\bibinfo{person}{Yang-Tian Sun}, \bibinfo{person}{Yi-Hua Huang}, \bibinfo{person}{Lin Ma}, \bibinfo{person}{Xiaoyang Lyu}, \bibinfo{person}{Yan-Pei Cao}, {and} \bibinfo{person}{Xiaojuan Qi}.} \bibinfo{year}{2024}\natexlab{}.
\newblock \showarticletitle{Splatter a Video: Video Gaussian Representation for Versatile Processing}.
\newblock \bibinfo{journal}{\emph{arXiv preprint arXiv:2406.13870}} (\bibinfo{year}{2024}).
\newblock


\bibitem[Tang et~al\mbox{.}(2025)]%
        {tang2025lgm}
\bibfield{author}{\bibinfo{person}{Jiaxiang Tang}, \bibinfo{person}{Zhaoxi Chen}, \bibinfo{person}{Xiaokang Chen}, \bibinfo{person}{Tengfei Wang}, \bibinfo{person}{Gang Zeng}, {and} \bibinfo{person}{Ziwei Liu}.} \bibinfo{year}{2025}\natexlab{}.
\newblock \showarticletitle{Lgm: Large multi-view gaussian model for high-resolution 3d content creation}. In \bibinfo{booktitle}{\emph{European Conference on Computer Vision}}. Springer, \bibinfo{pages}{1--18}.
\newblock


\bibitem[Teng et~al\mbox{.}(2023)]%
        {teng2023drag}
\bibfield{author}{\bibinfo{person}{Yao Teng}, \bibinfo{person}{Enze Xie}, \bibinfo{person}{Yue Wu}, \bibinfo{person}{Haoyu Han}, \bibinfo{person}{Zhenguo Li}, {and} \bibinfo{person}{Xihui Liu}.} \bibinfo{year}{2023}\natexlab{}.
\newblock \showarticletitle{Drag-a-video: Non-rigid video editing with point-based interaction}.
\newblock \bibinfo{journal}{\emph{arXiv}} (\bibinfo{year}{2023}).
\newblock


\bibitem[Voleti et~al\mbox{.}(2025)]%
        {voleti2025sv3d}
\bibfield{author}{\bibinfo{person}{Vikram Voleti}, \bibinfo{person}{Chun-Han Yao}, \bibinfo{person}{Mark Boss}, \bibinfo{person}{Adam Letts}, \bibinfo{person}{David Pankratz}, \bibinfo{person}{Dmitry Tochilkin}, \bibinfo{person}{Christian Laforte}, \bibinfo{person}{Robin Rombach}, {and} \bibinfo{person}{Varun Jampani}.} \bibinfo{year}{2025}\natexlab{}.
\newblock \showarticletitle{Sv3d: Novel multi-view synthesis and 3d generation from a single image using latent video diffusion}. In \bibinfo{booktitle}{\emph{European Conference on Computer Vision}}. Springer, \bibinfo{pages}{439--457}.
\newblock


\bibitem[Wang et~al\mbox{.}(2022)]%
        {wang2022pretraining}
\bibfield{author}{\bibinfo{person}{Tengfei Wang}, \bibinfo{person}{Ting Zhang}, \bibinfo{person}{Bo Zhang}, \bibinfo{person}{Hao Ouyang}, \bibinfo{person}{Dong Chen}, \bibinfo{person}{Qifeng Chen}, {and} \bibinfo{person}{Fang Wen}.} \bibinfo{year}{2022}\natexlab{}.
\newblock \showarticletitle{Pretraining is All You Need for Image-to-Image Translation}. In \bibinfo{booktitle}{\emph{arXiv}}.
\newblock


\bibitem[Wang et~al\mbox{.}(2025)]%
        {wang2024phidias}
\bibfield{author}{\bibinfo{person}{Zhenwei Wang}, \bibinfo{person}{Tengfei Wang}, \bibinfo{person}{Zexin He}, \bibinfo{person}{Gerhard Hancke}, \bibinfo{person}{Ziwei Liu}, {and} \bibinfo{person}{Rynson~WH Lau}.} \bibinfo{year}{2025}\natexlab{}.
\newblock \showarticletitle{Phidias: A generative model for creating 3d content from text, image, and 3d conditions with reference-augmented diffusion}.
\newblock \bibinfo{journal}{\emph{ICLR}} (\bibinfo{year}{2025}).
\newblock


\bibitem[Wu et~al\mbox{.}(2024b)]%
        {wu20244d}
\bibfield{author}{\bibinfo{person}{Guanjun Wu}, \bibinfo{person}{Taoran Yi}, \bibinfo{person}{Jiemin Fang}, \bibinfo{person}{Lingxi Xie}, \bibinfo{person}{Xiaopeng Zhang}, \bibinfo{person}{Wei Wei}, \bibinfo{person}{Wenyu Liu}, \bibinfo{person}{Qi Tian}, {and} \bibinfo{person}{Xinggang Wang}.} \bibinfo{year}{2024}\natexlab{b}.
\newblock \showarticletitle{4d gaussian splatting for real-time dynamic scene rendering}. In \bibinfo{booktitle}{\emph{Proceedings of the IEEE/CVF Conference on Computer Vision and Pattern Recognition}}. \bibinfo{pages}{20310--20320}.
\newblock


\bibitem[Wu et~al\mbox{.}(2023)]%
        {wu2023tune}
\bibfield{author}{\bibinfo{person}{Jay~Zhangjie Wu}, \bibinfo{person}{Yixiao Ge}, \bibinfo{person}{Xintao Wang}, \bibinfo{person}{Stan~Weixian Lei}, \bibinfo{person}{Yuchao Gu}, \bibinfo{person}{Yufei Shi}, \bibinfo{person}{Wynne Hsu}, \bibinfo{person}{Ying Shan}, \bibinfo{person}{Xiaohu Qie}, {and} \bibinfo{person}{Mike~Zheng Shou}.} \bibinfo{year}{2023}\natexlab{}.
\newblock \showarticletitle{Tune-a-video: One-shot tuning of image diffusion models for text-to-video generation}. In \bibinfo{booktitle}{\emph{Proceedings of the IEEE/CVF International Conference on Computer Vision}}. \bibinfo{pages}{7623--7633}.
\newblock


\bibitem[Wu et~al\mbox{.}(2024a)]%
        {wu2024cat4d}
\bibfield{author}{\bibinfo{person}{Rundi Wu}, \bibinfo{person}{Ruiqi Gao}, \bibinfo{person}{Ben Poole}, \bibinfo{person}{Alex Trevithick}, \bibinfo{person}{Changxi Zheng}, \bibinfo{person}{Jonathan~T Barron}, {and} \bibinfo{person}{Aleksander Holynski}.} \bibinfo{year}{2024}\natexlab{a}.
\newblock \showarticletitle{Cat4d: Create anything in 4d with multi-view video diffusion models}.
\newblock \bibinfo{journal}{\emph{arXiv preprint arXiv:2411.18613}} (\bibinfo{year}{2024}).
\newblock


\bibitem[Xie et~al\mbox{.}(2024)]%
        {xie2024sv4d}
\bibfield{author}{\bibinfo{person}{Yiming Xie}, \bibinfo{person}{Chun-Han Yao}, \bibinfo{person}{Vikram Voleti}, \bibinfo{person}{Huaizu Jiang}, {and} \bibinfo{person}{Varun Jampani}.} \bibinfo{year}{2024}\natexlab{}.
\newblock \showarticletitle{Sv4d: Dynamic 3d content generation with multi-frame and multi-view consistency}.
\newblock \bibinfo{journal}{\emph{arXiv preprint arXiv:2407.17470}} (\bibinfo{year}{2024}).
\newblock


\bibitem[Xu et~al\mbox{.}(2018)]%
        {xu2018youtube}
\bibfield{author}{\bibinfo{person}{Ning Xu}, \bibinfo{person}{Linjie Yang}, \bibinfo{person}{Yuchen Fan}, \bibinfo{person}{Dingcheng Yue}, \bibinfo{person}{Yuchen Liang}, \bibinfo{person}{Jianchao Yang}, {and} \bibinfo{person}{Thomas Huang}.} \bibinfo{year}{2018}\natexlab{}.
\newblock \showarticletitle{Youtube-vos: A large-scale video object segmentation benchmark}.
\newblock \bibinfo{journal}{\emph{arXiv preprint arXiv:1809.03327}} (\bibinfo{year}{2018}).
\newblock


\bibitem[Yang et~al\mbox{.}(2024c)]%
        {yang2024depth}
\bibfield{author}{\bibinfo{person}{Lihe Yang}, \bibinfo{person}{Bingyi Kang}, \bibinfo{person}{Zilong Huang}, \bibinfo{person}{Zhen Zhao}, \bibinfo{person}{Xiaogang Xu}, \bibinfo{person}{Jiashi Feng}, {and} \bibinfo{person}{Hengshuang Zhao}.} \bibinfo{year}{2024}\natexlab{c}.
\newblock \showarticletitle{Depth Anything V2}.
\newblock \bibinfo{journal}{\emph{arXiv preprint}} (\bibinfo{year}{2024}).
\newblock


\bibitem[Yang et~al\mbox{.}(2024b)]%
        {yang2024direct}
\bibfield{author}{\bibinfo{person}{Shiyuan Yang}, \bibinfo{person}{Liang Hou}, \bibinfo{person}{Haibin Huang}, \bibinfo{person}{Chongyang Ma}, \bibinfo{person}{Pengfei Wan}, \bibinfo{person}{Di Zhang}, \bibinfo{person}{Xiaodong Chen}, {and} \bibinfo{person}{Jing Liao}.} \bibinfo{year}{2024}\natexlab{b}.
\newblock \showarticletitle{Direct-a-video: Customized video generation with user-directed camera movement and object motion}. In \bibinfo{booktitle}{\emph{ACM SIGGRAPH 2024 Conference Papers}}. \bibinfo{pages}{1--12}.
\newblock


\bibitem[Yang et~al\mbox{.}(2024a)]%
        {yang2024deformable}
\bibfield{author}{\bibinfo{person}{Ziyi Yang}, \bibinfo{person}{Xinyu Gao}, \bibinfo{person}{Wen Zhou}, \bibinfo{person}{Shaohui Jiao}, \bibinfo{person}{Yuqing Zhang}, {and} \bibinfo{person}{Xiaogang Jin}.} \bibinfo{year}{2024}\natexlab{a}.
\newblock \showarticletitle{Deformable 3d gaussians for high-fidelity monocular dynamic scene reconstruction}. In \bibinfo{booktitle}{\emph{Proceedings of the IEEE/CVF Conference on Computer Vision and Pattern Recognition}}.
\newblock


\bibitem[Yenphraphai et~al\mbox{.}(2024)]%
        {yenphraphai2024image}
\bibfield{author}{\bibinfo{person}{Jiraphon Yenphraphai}, \bibinfo{person}{Xichen Pan}, \bibinfo{person}{Sainan Liu}, \bibinfo{person}{Daniele Panozzo}, {and} \bibinfo{person}{Saining Xie}.} \bibinfo{year}{2024}\natexlab{}.
\newblock \showarticletitle{Image sculpting: Precise object editing with 3d geometry control}. In \bibinfo{booktitle}{\emph{Proceedings of the IEEE/CVF Conference on Computer Vision and Pattern Recognition}}. \bibinfo{pages}{4241--4251}.
\newblock


\bibitem[Zhang et~al\mbox{.}(2023a)]%
        {zhang2023adding}
\bibfield{author}{\bibinfo{person}{Lvmin Zhang}, \bibinfo{person}{Anyi Rao}, {and} \bibinfo{person}{Maneesh Agrawala}.} \bibinfo{year}{2023}\natexlab{a}.
\newblock \showarticletitle{Adding conditional control to text-to-image diffusion models}. In \bibinfo{booktitle}{\emph{Proceedings of the IEEE/CVF International Conference on Computer Vision}}. \bibinfo{pages}{3836--3847}.
\newblock


\bibitem[Zhang et~al\mbox{.}(2023b)]%
        {zhang2023controlvideo}
\bibfield{author}{\bibinfo{person}{Yabo Zhang}, \bibinfo{person}{Yuxiang Wei}, \bibinfo{person}{Dongsheng Jiang}, \bibinfo{person}{Xiaopeng Zhang}, \bibinfo{person}{Wangmeng Zuo}, {and} \bibinfo{person}{Qi Tian}.} \bibinfo{year}{2023}\natexlab{b}.
\newblock \showarticletitle{Controlvideo: Training-free controllable text-to-video generation}.
\newblock \bibinfo{journal}{\emph{arXiv preprint arXiv:2305.13077}} (\bibinfo{year}{2023}).
\newblock


\end{thebibliography}

\end{document}